\documentclass[10pt,journal,compsoc]{IEEEtran}
\usepackage[utf8]{inputenc}
\ifCLASSOPTIONcompsoc
  \usepackage[nocompress]{cite}
\else
  \usepackage{cite}
\fi

\ifCLASSINFOpdf
\else
\fi

\usepackage{times}
\usepackage{epsfig}
\usepackage{graphicx}
\usepackage{amsmath}
\usepackage{amssymb}
\usepackage{gensymb}

\usepackage{paralist}
\usepackage{amsthm}
\usepackage{color}
\usepackage[normalem]{ulem}
\usepackage{booktabs}
\usepackage{multirow}
\usepackage[misc,geometry]{ifsym}
\usepackage{stmaryrd}
\usepackage{algorithm}
\usepackage{algorithmic}
\usepackage{array}
\usepackage{url}
\usepackage{subfigure}
\usepackage{mathtools,nccmath}
\usepackage[hang,flushmargin]{footmisc}
\usepackage[pagebackref=false,breaklinks=true,letterpaper=true,colorlinks,citecolor=blue,linkcolor=blue,bookmarks=false]{hyperref}
\usepackage{comment}

\newcommand{\balpha}{\mbox{\boldmath $\alpha$}}

\newcommand{\be}{\begin{eqnarray}}
\newcommand{\ee}{\end{eqnarray}}
\newcommand{\bee}{\begin{eqnarray*}}
\newcommand{\eee}{\end{eqnarray*}}

\newcommand{\matrixb}{\left[ \begin{array}}
\newcommand{\matrixe}{\end{array} \right]}

\makeatletter
\usepackage{xspace}
\DeclareRobustCommand\onedot{\futurelet\@let@token\@onedot}
\def\@onedot{\ifx\@let@token.\else.\null\fi\xspace}

\def\eg{\emph{e.g}\onedot} 
\def\ie{\emph{i.e}\onedot}

\def\etal{\emph{et al}\onedot}
\makeatother

\newcommand{\Tref}[1]{Table~\ref{#1}}

\newcommand{\Fref}[1]{Figure~\ref{#1}}

\newcommand{\Cref}[1]{Chap.~\ref{#1}}
\newcommand{\Sref}[1]{Section~\ref{#1}}

\newcommand{\R}{\mathbb{R}}

\newcommand{\paragrp}[1]{\vspace{1mm}\noindent\textbf{#1}\,\,\,}

\newcommand{\ra}[1]{\renewcommand{\arraystretch}{#1}}

\begin{document}
%
\title{
Learning to Localize Sound Sources\\
in Visual Scenes: Analysis and Applications
}
%
%
%
%

\author{Arda Senocak, 
		Tae-Hyun Oh, 
		Junsik Kim, 
        Ming-Hsuan Yang, 
		and~In So Kweon 
\IEEEcompsocitemizethanks{
\IEEEcompsocthanksitem A. Senocak, J. Kim and I. S. Kweon are with School of Electrical Engineering, KAIST, Daejeon, Republic of Korea.
\IEEEcompsocthanksitem T.-H. Oh is with MIT CSAIL, Cambridge, MA, USA.
\IEEEcompsocthanksitem Ming-Hsuan Yang is with Dept. of Electrical Engineering and Computer Science at University of California, Merced.
\IEEEcompsocthanksitem Corresponding authors: T.-H. Oh (\texttt{taehyun@csail.mit.edu}) and I.S. Kweon (\texttt{iskweon@kaist.ac.kr}).
}
}



\IEEEtitleabstractindextext{%
\begin{abstract}
Visual events are usually accompanied by sounds in our daily lives. 
However, can the machines learn to correlate the visual scene and sound, as well as localize the sound source only by observing them like humans?
To investigate its empirical learnability, in this work we first present a novel unsupervised algorithm to address the problem of localizing sound sources in visual scenes. 
In order to achieve this goal, a two-stream network structure which handles each modality with attention mechanism is developed for sound source localization. 
The network naturally reveals the localized response in the scene
without human annotation. 
In addition, a new sound source dataset is developed for performance evaluation.
Nevertheless, our empirical evaluation shows that the unsupervised method generates false conclusions in some cases. Thereby, we show that this
false conclusion cannot be fixed without human prior knowledge due to the well-known correlation and causality mismatch misconception. 
To fix this issue, we extend our network to the supervised and semi-supervised network settings via a simple modification due to the general architecture of our two-stream network. 
We show that the false conclusions can be effectively corrected even with a small amount of supervision, \ie, semi-supervised setup. 
Furthermore, we present the versatility of the learned audio and visual embeddings on the cross-modal content alignment and we extend this proposed algorithm to {a new application,} sound saliency based automatic camera view panning in 360\degree~videos.
\end{abstract}

}

\maketitle
\IEEEpeerreviewmaketitle


\section{Introduction}\label{sec:intro}

\begin{figure}[t]
\centering

		\includegraphics[width=.7\linewidth]{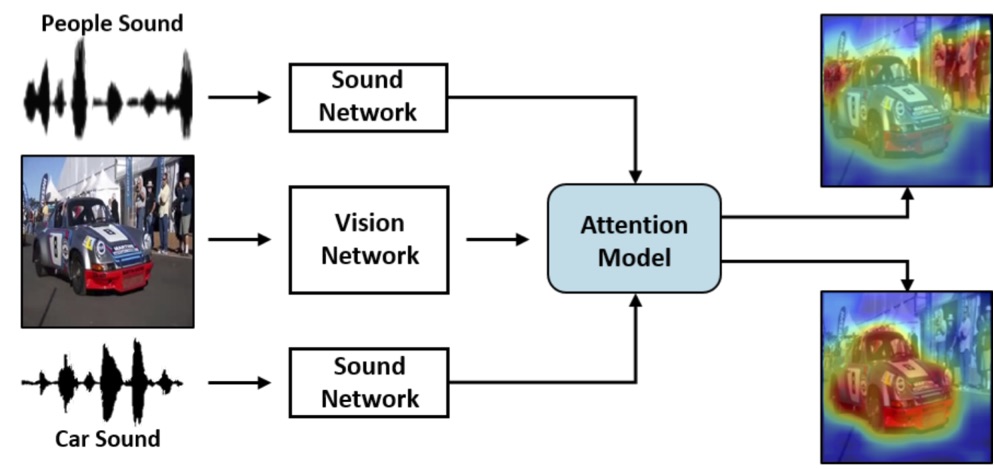}

	\caption{\textbf{Where do these sounds come from?} We show an example of interactive sound source localization by the proposed algorithm. In this work, we demonstrate how to learn to localize {sound} sources (objects) from the sound signals in visual scenes.}
\label{fig:first_page_fig}
\vspace{-3mm}
\end{figure}

\IEEEPARstart{U}{}nderstanding the world that surrounds us is a multi-modal experience. 
We perceive the world by using multiple senses at the same time. 
Visual events are typically associated with sounds and they are often integrated. 
When we see that a car is moving, we hear the engine sound at the same time, \ie, co-occurrence.  
Sound carries rich information regarding the spatial and temporal cues of the source within a visual scene.
As shown in the bottom example of \Fref{fig:first_page_fig}, the engine sound suggests where the source may be in the physical world~\cite{gaver1993world}.
This implies that sound is not only complementary to the visual information, but also {correlated} to visual events.

{Human perception is also multi-modal.} Humans observe tremendous amount of combined audio-visual data and learn the correlation between them throughout their whole life unconsciously~\cite{gaver1993world}. {From this life-long experiences,} humans can
understand the object or event that causes sound, and localize the sound source even without separate education.
Naturally, videos and their corresponding sounds also occur together in a synchronized way. 
When considering an analogous behavior in the context of machine learning, the following question may arise:
given a plenty of video and sound clip pairs, can a machine learning model learn to associate the sound with {the} visual scene to reveal the sound source location without any supervision similar to human perception?
This question is the motivation of our work.

\begin{figure*}
	\centering

		\includegraphics[width=.8\linewidth]{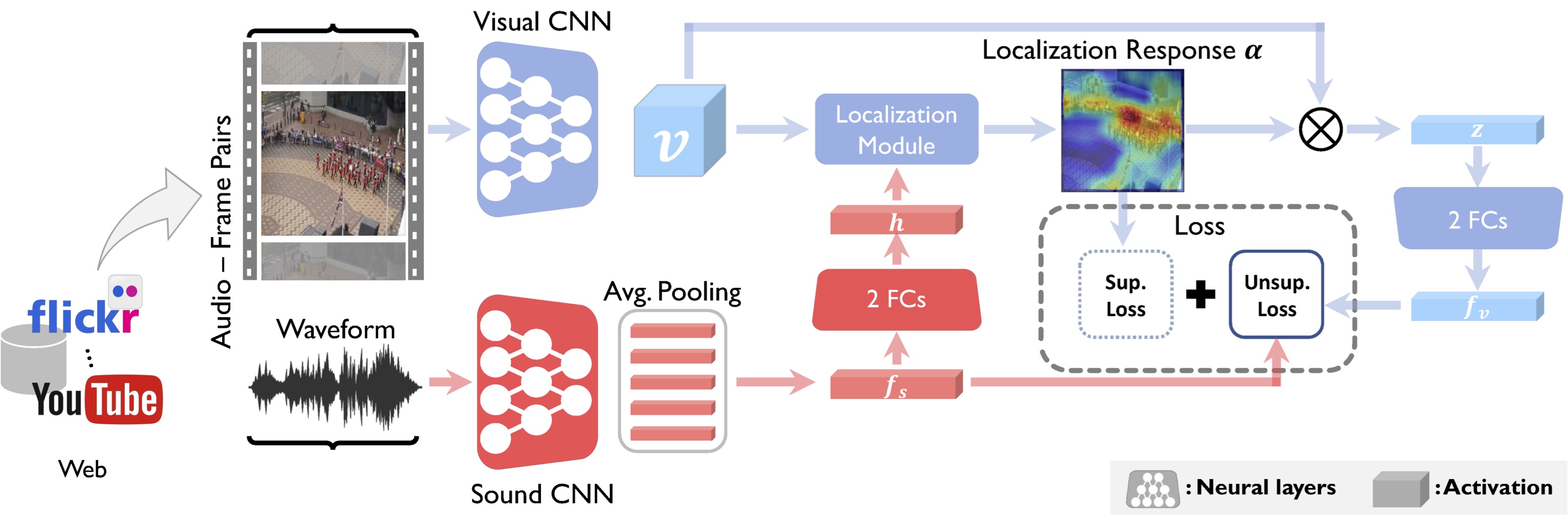}
	\caption{\textbf{Network architecture.} This architecture is designed to tackle the problem of sound source localization with self-supervised learning. The network uses video frame and sound pairs to learn to localize the sound sources. Each modality is processed in its own network. After integrating (correlating) the information from the sound context vector $\mathbf{h}$ and the activations of {the} visual network, the localization module (attention mechanism) localizes the sound source. By adding {the} supervised loss component into this architecture, it can be converted to a unified architecture which can work as supervised or semi-supervised learning as well. In the figure, FC stands for the fully connected layer, and $\bigotimes$ denotes the weighted sum pooling across spatial dimensions.}
	\label{fig:pipeline}
\end{figure*}

There has been significant progress in the field of audio-visual learning recently by the advances of deep learning~\cite{senocak2018learning,arandjelovic2018objects,gao2018learning,Zhao_2018_ECCV,ephrat2018looking,harwath2018jointly,owens2018audioECCV,tian2018ave,Zhou2018visual2sound,Kim2018on}.
In this work, we specifically focus on whether a neural model can learn to extract the spatial correspondence between visual and audio information by simply watching and listening to videos in a self-supervised way, \ie, learning based sound source localization.
To this end, we design a two-stream network architecture (sound and visual networks), where each network facilitates each modality, and a localization module which incorporates the attention mechanism as illustrated in \Fref{fig:pipeline}.
The proposed network is designed to leverage the co-occurrence of both modalities, visual appearance of a sound source object and its sound, without supervision, \ie, self-supervision.

The learning task for sound source localization from listening is challenging, especially from unlabeled data. 
From our experiments with the proposed self-supervised model, we observe a classical phenomenon~\cite{shalev2014understanding} in learning theory, \ie, pigeon superstition,
which describes a learning model is biased to conclude the resulting localization to be semantically unmatched in our case.
{We show that it is difficult for unsupervised learning methods to disambiguate  sound sources purely based on correlations 
from a static single image and mono-channel audio
without some supervisory signals.}

We correct this issue by providing a small amount of supervision in a semi-supervised setting.
By virtue of our unified architecture design, we can easily transform our network to the self-supervised, fully-supervised, or semi-supervised framework by simply adding a supervised loss, depending on the availability of annotated data.
This allows us to resolve the aforementioned issue.
To incorporate in the unified architecture and to evaluate the proposed methods, we annotate a new sound source localization dataset.
To the best of our knowledge, this dataset is the first to address the problem of learning based sound localization. 

The contributions of this work are summarized as follows:

{

	\begin{itemize}
		\item We introduce a learning framework to localize the sound source using an attention mechanism, which is guided by sound information, from a paired sound and video frame.
		The sound source localization can be interactive with given sound input.
		\item We propose a unified end-to-end deep convolutional neural network architecture that accommodates unsupervised, semi-supervised, and fully-supervised learning.
		\item We collect and annotate a new sound source localization dataset, which {provides supervised information and facilitates quantitative and qualitative analysis}.
        \item We quantitatively and qualitatively demonstrate that the learning based sound source localization is not correctly solved
        with a purely unsupervised approach, but can be fixed even with a small amount of supervisory data. 
        \item We further analyze that learned embeddings are able to represent the semantic context in cross-domain samples.
        \item We present a new application of sound guided saliency prediction for 360$\degree$~videos and 360$\degree$~content exploration.
	\end{itemize}
}
\vspace{-1mm}\section{Related Work and Problem Context}
\label{sec:related}
Cross-modality signals have been used as supervisory information for numerous tasks.
The recent years have witnessed significant progress in understanding the correlation between sound and visual signals.
To put this work into proper context, we review recent methods on joint audio-vision models, sound source localization, and attention mechanisms.

\paragrp{Audio-visual representation learning.}
Visual scenes in real-world accompany sound in many cases.
This co-occurrence property of the two modalities has been recently exploited by Owens~\etal\cite{Ambient,owens2018learning} and Aytar~\etal\cite{SoundNet}, where Owens~\etal use sound as supervisory signals by virtue of its natural synchronization with visual input, while Aytar~\etal regard visual imagery as supervision for sound.
Both methods learn {the} representation of one of the modalities while using the other one as supervision, \ie, transferring knowledge.
On the other hand, Arandjelovi\'c~\etal~\cite{Arandjelovic17} learn audio and visual representations by using both modalities in an unsupervised manner. 
Aytar~\etal\cite{aytar2017see} also explore aligned representations by adding another modality, text. 
All the above-mentioned methods use a static image and corresponding audio pair. 
In contrast, Owens~\etal\cite{owens2018audioECCV} and Korbar~\etal\cite{korbar2018Cooperative} analyze audio-visual \emph{actions} and learn representations by using videos. 

\paragrp{Sound source localization in visual scenes.}
Prior to the recent advances of deep learning, computational methods for sound source localization rely on synchrony~\cite{Synchrony} of low-level features of sounds and videos (\eg, raw waveform signals and intensity values respectively), spatial sparsity prior of audio-visual events~\cite{PixelSound}, low-dimensionality~\cite{fisher2001learning}, hand-crafted motion cues, and segmentation~\cite{Harmony,Motion}. 
In contrast, the proposed network is developed in an unsupervised manner by only watching and listening to videos 
{without using any hand-designed rules such as the ones mentioned above.}
Furthermore, our semi-supervised architecture does not require hand-crafted prior knowledge except {for} a small amount of annotated data. 

Acoustic hardware based approaches~\cite{van2002optimum,Zunino} have been practically used in surveillance and instrumentation engineering. 
These methods require specific devices, \eg, microphone arrays, to capture phase differences of sound arrival.
In this work, we learn sound source localization in the visual domain without any special devices but a microphone to capture sound. 
We restrict to mono channel sound and focus on the relationship between visual \emph{context} and sound, rather than other physical relationships, \eg, phase differences of sound arrival or {motions}.

With new advances in deep learning, this task has attracted more attention~\cite{senocak2018learning,Zhao_2018_ECCV,arandjelovic2018objects,owens2018audioECCV,owens2018learning,tian2018ave}. 
Since the approach we propose recently~\cite{senocak2018learning}, some interesting methods on the sound source localization task have been developed. 
Although Arandjelovi\'c~\etal~\cite{Arandjelovic17} show that activation maps can be used to localize objects, the localization results are solely from examining the units in the vision subnetwork. 
This work is further extended to locate the objects based on  sound sources~\cite{arandjelovic2018objects}. 
While this method largely focuses on localizing musical instruments and their sounds, our method is designed for generic scenes. 
Furthermore,  our  networks  have  an  attention layer  that interacts  between  the  two  modalities  and  reveals  the localization  information  of  the  sound  source.
In \cite{Zhao_2018_ECCV}, Zhao~\etal also explore the sound source localization in the musical instruments domain. 
On the other hand, several  methods~\cite{owens2018audioECCV,korbar2018Cooperative} are designed to localize actions in videos, rather than objects in static images with an unsupervised learning method. 
Tian~\etal~\cite{tian2018ave} also focus 
on audio-visual event {localization} but with fully and weakly-supervised approaches.
Recently, Harwath~\etal develop a method for grounding spoken words in images~\cite{harwath2018jointly}.

In the context of sound source separation,
we note several methods~\cite{ephrat2018looking, Afouras18, gao2018learning, owens2018audioECCV}
demonstrate that visual information plays an important {role in} such tasks. 
Nevertheless, the goals of these methods are different from the focus of this work.

\paragrp{Sound source localization in psychophysics.}
Our work is motivated by the findings in psychology and cognitive science on the sound source localization capability of humans~\cite{gaver1993world,jones1975eye,Majdak2010,Shelton1980,Bolia,Perrott}.
Gaver~\etal\cite{gaver1993world} study how humans learn about objects and events from sound in everyday listening.
This study elucidates how humans can find the relationship between visual and sound domains in an event centric view.
Numerous methods in this line analyze the relationship between visual information and sound localization.
These findings show that visual information correlated to sound improves the efficiency of search~\cite{jones1975eye} and accuracy of localization~\cite{Shelton1980}.
Recent methods~\cite{Majdak2010,Bolia,Perrott} extend the findings of human performance on sound source localization against visual information in 3D space. 
These studies evidently show that sound source localization capability of humans is guided by visual information, and two sources of information are closely correlated that humans can unwittingly learn such capability.

\paragrp{Visual and aural modality association.}
Inspired by the human vision system~\cite{corbetta2002control}, numerous  attention models~\cite{show,zhou2016learning} have been developed for vision tasks. 
We extend the use of the computational attention mechanism to multisensory integration, 
in that 
the sound localization behavior in imagery resembles human attention.
The existence of multisensory integration including visual and auditory stimuli in our human brain and working sites are known
as superior colliculi~\cite{stein2008multisensory}.
In this work, we adopt {a} similar principle with the attention mechanism in \cite{show} to enable our networks to interact with sound context and visual representation across spatial axes.
\section{Proposed Algorithm}
\label{sec:approach}
We first present a neural network
to address the problem of vision based sound localization
within the unsupervised learning framework. 
Next, we show that it can be extended to supervised and semi-supervised frameworks by simply appending a supervised loss.
To deal with cross-modality signals from sounds and videos, we use a two-stream network architecture.
The network consists of three main modules: sound network, visual network and attention model as illustrated in  
\Fref{fig:pipeline}.
\subsection{Sound Network}\label{sec:soundnet}\vspace{-1mm}
In this work, we focus on learning the semantic relationship between the mono channel audio and a single frame without taking motion into consideration. 
Thus, it is important to capture the context of sound rather than catching low-level signals~\cite{gaver1993world}.
In addition, sound signals are 1-D with varying temporal length.
We encode sound signals into high-level concepts by using the convolutional module (\texttt{conv}), rectified linear unit (\texttt{ReLu}) and pooling (\texttt{pool}), and those stacking layers~\cite{zeiler2014visualizing}.
We use a 1-D deep convolutional architecture which is invariant to input length due to the fully convolutional feature via the use of global average pooling over sliding windows. 

The proposed sound network consists of 10 layers and takes raw waveform as input.  
The first  \texttt{conv} layers (up to \texttt{conv8}) are similar to the SoundNet~\cite{SoundNet}, 
but with $1000$ filters followed by global average pooling across the temporal axis within a sliding window (\textit{e.g}, 20 seconds in this work).
The global average pooling facilitates to handle variable length inputs to be a fixed dimension vector~\cite{van2013deep}, \ie, the output activation of \texttt{conv8} followed by the average pooling is always a single $1000$-dimensional vector.
We denote this sound representation after the average pooling as $\mathbf{f}_s$.

To capture high level concept of sound signals, the 9-th and 10-th layers consist of ReLU followed by fully connected (\texttt{FC}) layers.
The output of the 10-th \texttt{FC} layer (\texttt{FC10}) is $512$-dimensional, and is denoted as $\mathbf{h}$. 
We use $\mathbf{h}$ to interact with features from the visual network, and induce $\mathbf{h}$ to resemble visual concepts.
Among these two features, we note $\mathbf{f}_s$ preserves more sound concept while $\mathbf{h}$ captures correlated information related to visual signals. 

\subsection{Visual Network}
\label{sec:vis_model}
The visual network is composed of the image feature extractor and the localization module. 
To extract features from visual signals, we use 
an architecture similar to the VGG-16 model~\cite{Simonyan15} up to \texttt{conv5\_3} and feed a color video frame of size $H{\times} W$ as input.
	We denote the activation of \texttt{conv5\_3} as ${\mathcal{V}} \in \mathbb{R}^{H' {\times} W' {\times} D}$, where $H'{=}\lfloor \tfrac{H}{16} \rfloor$, $W'{=}\lfloor \tfrac{W}{16} \rfloor$ and $D=512$.
	Each $512$-D activation vector from \texttt{conv5\_3} contains local visual context information, and spatial information is  preserved in the $H' {\times} W'$ grid.

In our model, the activation $\mathcal{V}$ interacts with the sound embedding $\mathbf{h}$ for revealing sound source location information in the grid, which is denoted as the localization module (Section~\ref{sec:att_model}). 
This localization module returns a confidence map of {the} sound source and a representative visual feature vector $\mathbf{z}$ corresponding to {the} estimated location of the given input sound.
The visual feature $\mathbf{z}$ is passed through two $\{$\texttt{ReLu-FC}$\}$ blocks to compute the visual embedding $\mathbf{f}_v$, which is the final output of the visual network.

\subsection{Localization Network}\label{sec:att_model}
Given extracted visual and sound concepts, 
the localization network generates the sound source location.
We compute a soft confidence score map as a sound source location representation.
This may be modeled based on the attention mechanism in the human visual system~\cite{
corbetta2002control}, where according to given conditional information, related salient features are dynamically and selectively brought out to the foreground.
This motivates us to exploit the neural attention mechanism~\cite{show,bahdanau2014neural} in our context.

For simplicity, instead of using a tensor representation for the visual activation ${\mathcal{V}} \in \mathbb{R}^{H' {\times} W' {\times} D}$, we denote the visual activation as a reshaped matrix form $\mathbf{V}\,{=}\,[\mathbf{v}_1;\cdots;\mathbf{v}_M]\in\R^{M\times D}$, where $M = H'W'$.
For each location $i\in\{ 1{,}{\cdots}{,}M \}$, the attention mechanism $g_{\mathtt{att}}$ generates the positive weight $\alpha_i$ by the interaction between the given sound embedding $\mathbf{h}$ and $\mathbf{v}_i$, where $\alpha_i$ is the attention measure. 
The attention $\alpha_i$ can be interpreted as the probability that the grid $i$ is likely to be the right location related to the sound context, and computed by
\begin{equation}
	\alpha_i = \tfrac{\exp({a}_{i})}{\sum_j \exp({a}_{j})}, \quad\textrm{where}~a_i = g_\mathtt{att}(\mathbf{v}_i,\mathbf{h}),
	\label{eq:attention_softmax}
\end{equation}
where the normalization by the softmax is suggested by \cite{bahdanau2014neural}.

In contrast to the work~\cite{show,bahdanau2014neural} that uses a multi-layer perceptron as $g_\mathtt{att}$, we use the simple normalized inner product operation that does not require any learning parameter.
Furthermore, it is intuitively interpretable {as}
the operation measures the cosine similarity between two heterogeneous vectors, $\mathbf{v}_i$ and $\mathbf{h}$, \ie, correlation. We also propose an alternative attention mechanism to suppress negative correlation values as:
\begin{eqnarray}
&&\hspace{-8mm}\textrm{[Mechanism 1]}\quad g_\mathtt{cos}(\mathbf{v}_i, \mathbf{h}) = \bar{\mathbf{v}}_i^\top \bar{\mathbf{h}},\\
&&\hspace{-8mm}\textrm{[Mechanism 2]}\quad
g_\mathtt{ReLu}(\mathbf{v}_i, \mathbf{h}) = \max(\bar{\mathbf{v}}_i^\top \bar{\mathbf{h}}, 0),
\end{eqnarray}
where $\bar{\mathbf{x}}$ denotes a $\ell_2$-normalized vector. 
This is different from the mechanism proposed in~\cite{show,bahdanau2014neural,zhou2016learning}.
Zhou~\etal\cite{zhou2016learning} use a typical linear combination without normalization, and thus it can have an arbitrary range of values.
Both mechanisms in this work are based on the cosine similarity of the range $[-1,1]$.
The attention measure $\balpha$ computed by either mechanism  describes the sound and visual context interaction in a map.
To draw a connection to $\balpha$ with sound source location, similar to \cite{show,bahdanau2014neural}, 
we compute the representative context vector $\mathbf{z}$ that corresponds to the local visual feature at the sound source location.
Assuming that $\mathbf{z}$ is a stochastic random variable and $\balpha$ represents the sound source location reasonably well, 
we regard the attention locations $i$ as latent variables by parameterizing $p(i|\mathbf{h}) = \alpha_i$.
Then, the visual feature $\mathbf{z}$ can be obtained by 
\begin{equation}
\label{eq:context_vector}
\mathbf{z} = \mathbb{E}_{p(i|\mathbf{h})}[\hat z] = \sum\nolimits_{i=1}^M \alpha_i \mathbf{v}_i.
\end{equation}

As described in \Sref{sec:vis_model}, we transform a visual feature vector $\mathbf{z}$ to a visual representation $\mathbf{f}_v$.
We adapt $\mathbf{f}_v$ to be comparable with the sound features $\mathbf{f}_s$ obtained from the sound network, such that we learn the features to share embedding space.
During the learning phase, the back-propagation encourages $\mathbf{z}$ to be related to the sound context.
{Importantly, while $\mathbf{z}$ is parameterized by $\balpha$ and $\mathbf{v}$, since $\balpha$ is the only variable conditioned by the sound context, $\balpha$ is learned to adjust $\mathbf{z}$ in a way that it contains the sound context, \ie, learned to localize the sound.}

\section{Localizing Sound Source via Listening}\label{sec:learning}
Our learning model determines a video frame and audio signals are similar to each other or not at each spatial location. 
With the proposed two-stream network, we obtain predictions from each subnetwork for the frame and sound.
If the visual network considers that a given frame contains a motorcycle and the sound network also returns similar output, then the predictions of these networks are likely to be similar and close to each other in the feature space, and vice versa.
This provides valuable information for learning to localize sound sources in different settings. 

\paragrp{Unsupervised learning.}    
In the feature space, we impose that  $\mathbf{f}_v$ and $\mathbf{f}_s$ from the corresponding pairs (positive) are close to each other, while negative pairs are far from each other.
Using $\mathbf{f}_v$ from a video frame as a query, we obtain its positive pair by taking $\mathbf{f}_s$ from the sound wave of a sliding window around the video frame in the same video, and extract negative one from another random video. 
Given queries and those positive and negative pairs, we use the triplet loss~\cite{hoffer2015deep}.
The loss is designed to map the positive samples into the similar location as much as possible with the query in the feature space, while mapping the negative samples into distant locations. 

A triplet network computes two distance terms:
\begin{equation}
[d^+, d^-] \equiv [\|\mathbf{f}_v - \mathbf{f}_s^+\|_2, \|\mathbf{f}_v - \mathbf{f}_s^-\|_2] = T(\mathbf{f}_v,\mathbf{f}_s^-,\mathbf{f}_s^+),
\end{equation}
where $T(\cdot)$ denotes the triplet network, $(\mathbf{x}, \mathbf{x}^+, \mathbf{x}^-)$ represents a triplet of query, positive and negative samples. 
To impose the constraint $d^+ < d^-$, we use the distance ratio loss~\cite{hoffer2015deep}.  
The unsupervised loss function is defined as
\begin{equation}
	\mathcal{L}_{U}(D^+,D^-) = \left\| \left[D^+, D^-\right] - \left[0, 1 \right] \right\|^2,\vspace{-1mm}
	\label{eq:loss}
\end{equation}
where $D_{\pm} {=} \tfrac{\exp(d_\pm)}{\exp(d_+) + \exp(d_-)}$.
For the positive pair, the unsupervised loss imposes the visual feature $\mathbf{f}_v$ to resemble $\mathbf{f}_s$.
For the visual feature $\mathbf{z}$ to generate such $\mathbf{f}_v$, the weight $\balpha$ needs to select causal locations by the correlation between $\mathbf{h}$ and $\mathbf{v}$.
This results in $\mathbf{h}$ to share the embedding space with $\mathbf{v}$, and $\mathbf{f}_s$ also needs to encode the context information that correlates with {the} video frame.
This forms a cyclic loop, as shown in \Fref{fig:pipeline}, which allows to learn a shared representation that can be used for sound localization. 

\begin{figure}
	\centering

		\includegraphics[width=0.9\linewidth]{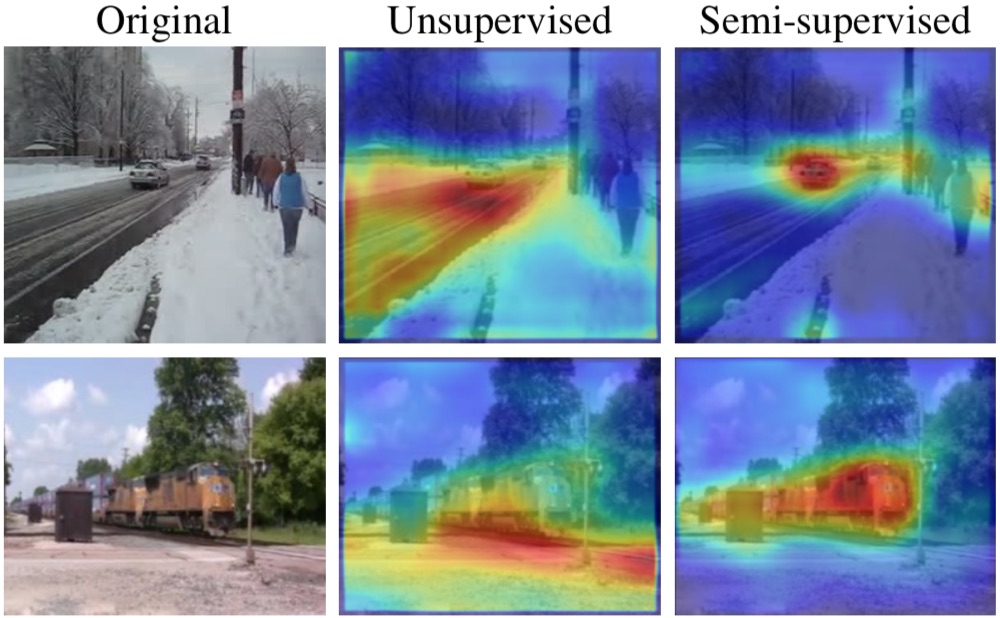}
\vspace{-3mm}
\caption{\textbf{Semantically unmatched results.} We show some cases where the proposed network with unsupervised learning draws false conclusions. 
We correct this issue by providing prior knowledge.}
\label{fig:pidgeon}
\end{figure}

Although the unsupervised learning method appears to perform well (in terms of the metric),  
we encounter some semantically unmatched results.
For example, as shown in Figure~\ref{fig:pidgeon}, even though we present a train sound with a train image, the proposed model localizes railways rather than the train.
This false conclusion by the model can be explained as follows.
In the early stage of the training, our model mistakenly concludes with false random output (\eg, activation on the road given {the} car sound). 
However, it obtains a good score (as the score is measured from  weak supervisions of corresponding pairs), thereby the model is trained to behave similarly for such scenes. 
Thus, the model reinforces itself to receive good scores in similar examples. 
As a specific example, 
in the road case of Figure~\ref{fig:pidgeon},
the proposed network consistently sees similar roads with car sounds during training, because cars are typically accompanied by roads. 
Since the road has a consistent appearance and typically occupies larger regions compared to diverse appearance of cars (or non-existence of any car in the frame at times), it is difficult for the model to discover a true causality relationship with the car without supervisory feedback. 
This ends up biasing toward a certain semantically unrelated output.

A similar phenomenon is often observed in the learning models~\cite{shalev2014understanding} and animals, which is known as the pigeon superstition phenomenon\footnote{It is an experiment~\cite{Skinner1948} about delivering food to hungry pigeons in a cage at regular time intervals regardless of the bird behavior.
When food was first delivered, it is found that each pigeon was engaging in some activity. 
Then they started doing the same action, believing that by acting in that way, food would arrive, \ie, reinforced to do a specific action.
	Such self-reinforcement occurs regardless of the truth of causality of the event and its chance.
	Some of such fundamental issues that naturally occur in the context of animal learning also appear in machine learning.}.
Since the relationship between source and result information was not trivial, the learner made a wrong decision with high confidence, in that there is
no way to validate and correct such a superstition for the learner with only unsupervised loss.
It has been known that, without directly related external prior knowledge, no further learning is possible~\cite{shalev2014understanding}.
While other types of prior knowledge would be an option, we provide a small amount of annotated data in the semi-supervised setting to address this issue (see the last column of Figure~\ref{fig:pidgeon}).

\paragrp{Semi-supervised learning.}
Even a small amount of prior knowledge can induce effective inductive bias.
We add a supervised loss to the proposed network architecture under the semi-supervised learning setting as
\begin{equation}
\label{eq:loss_semisup}
\begin{aligned}
&\hspace{-13mm}\mathcal{L}( \mathbf{f}_v, \mathbf{f}^+_s,\mathbf{f}^-_s,\balpha, \balpha_\mathtt{GT}) = \\
&\hspace{10mm}\mathcal{L}_{U}(\mathbf{f}_v, \mathbf{f}^+_s,\mathbf{f}^-_s) +
\lambda(\balpha_\mathtt{GT})\cdot \mathcal{L}_{S}(\balpha, \balpha_\mathtt{GT}),
\end{aligned}
\end{equation}
where $\mathcal{L}_{\{U, S\}}$ denote unsupervised and supervised losses respectively, $\balpha_\mathtt{GT}$ denotes the ground-truth (or reference) attention map, and $\lambda(\cdot)$ is a function for controlling the data supervision type.
The unsupervised loss $\mathcal{L}_{U}$ is 
same as \eqref{eq:loss}.
The supervised loss $\mathcal{L}_{S}$ 
is defined by
\begin{equation}
\label{eq:cross_entropy}
    \mathcal{L}_{S}(\balpha, \balpha_\mathtt{GT}) =  - \sum\nolimits_i {{\alpha_{\mathtt{GT},i}} \log({\alpha_i})},
\end{equation}
where $i$ denotes the location index of the attention map and $\alpha_{\mathtt{GT},i}$ is a binary value. 
{The cross entropy loss  is selected as empirically it performs slightly better than other functions.}
We set $\lambda(\mathbf{x}) = 0$ if $\mathbf{x} \in \emptyset$, or otherwise $1$.
With this formulation, we can easily adapt the loss to be either supervised or unsupervised one according to the existence of $\balpha_\mathtt{GT}$ for each sample.
In addition, \eqref{eq:loss_semisup} can be directly utilized for fully supervised training.

\begin{figure}[t]
\centering

		\includegraphics[width=\linewidth]{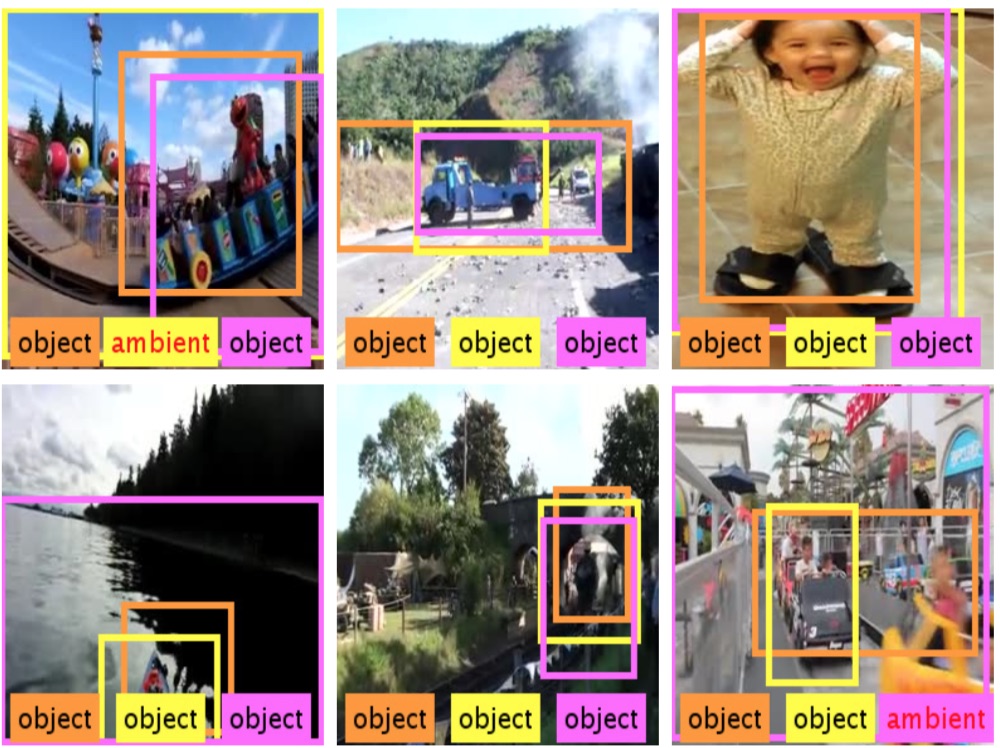}
\vspace{-3mm}
\caption{\textbf{Sound source localization dataset.} The location and type of the sound sources (object vs.\ non-object/ambient) are annotated. 
This dataset is used for testing how well our network learns the sound localization and  for providing supervision to the unified architecture.}
\label{fig:dataset}
\end{figure}
\section{Experimental Results}
\label{sec:experiments}
For evaluation, we first construct a new sound source localization dataset which facilitates quantitative and qualitative evaluation.
In this section, we discuss our empirical observations, and demonstrate how such issues can be corrected with a small amount of supervision.
In addition, we evaluate the unified network in unsupervised, semi-supervised and supervised learning schemes.
We implement our architecture with TensorFlow~\cite{tensorflow}.
For training, we use ADAM~\cite{AdamOpt} optimizer with the fixed learning rate of 0.0001, and a batch size of 30. For the visual CNN, while the architecture supports any resolution of input size due to the fully convolutional design, we resize the input frame of $320\times 320$ pixels as input during training.

\begin{figure*}
\vspace{-1mm}

		\includegraphics[width=\linewidth]{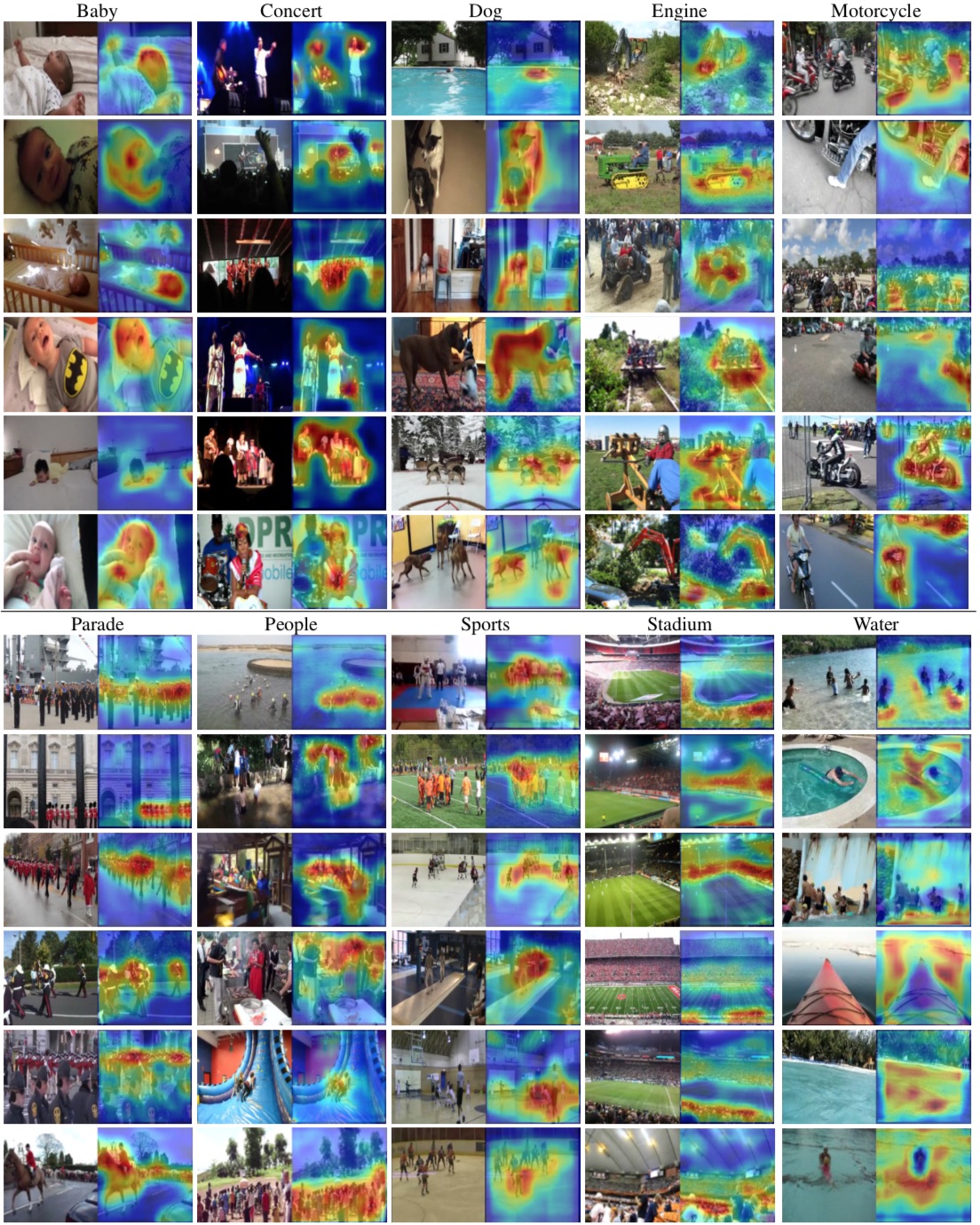}
\vspace{-2mm}
   \caption{\textbf{Qualitative sound localization results from unsupervised network.}
We feed image and sound pairs through our unsupervised network to localize sound sources.
Titles of the columns {are subjective annotations of contents in the corresponding sounds} and they are shown only for visualization purpose to give an idea about the sound context to readers: We do not use explicit labels.}
\vspace{-2.5mm}
\label{fig:heatmap}
\end{figure*}
\vspace{-1mm}
\subsection{Dataset}\vspace{-1mm}
\label{sec:dataset}
In order to train our network to localize the sound sources, we leverage the unlabeled Flickr-SoundNet~\cite{SoundNet,Yahoo} dataset, which consists of more than two million unconstrained sound and image pairs. 
We use a random subset of this dataset (144k pairs) to train our network.

For performance evaluation, we collect a new dataset that sources of sounds are annotated in image coordinates using pairs from the Flickr-SoundNet set.
This dataset not only facilitates quantitative and qualitative evaluation, but also provides annotations for training supervised models.
We randomly sample 5k frames and corresponding sounds from Flickr-SoundNet.
Three subjects independently annotate sound source by giving generic instructions as follows: 1) listen 20 seconds of sound and draw a bounding box on the frame at the regions where the dominant sound would come from, and 2) tag the bounding box as \emph{object} or \emph{ambient}.

Since the dataset we use contains unconstrained videos, some frames do not have the sound source in the frame or it cannot be represented by drawing a bounding box, \eg, wind sound. 
The tag is used to distinguish this case as ``object'' or ``ambient/not object'' for each bounding box.
After the annotation process, we filter out ``ambient/not object'' image-sound pairs. 
{Among the remaining pairs, we select the ones that all subjects agree with the sound indicating objects present in the frame.}
As such, we have a supervised set of $2,786$ pairs.
\Fref{fig:dataset} shows some sample images. 

\subsection{Results and Analysis}\vspace{-1mm}
We introduce a metric for quantitative performance evaluation of sound localization. 

\paragrp{Evaluation metrics.}
We have three annotations from three subjects for each data point. 
As some examples could be ambiguous, \eg, the left and right examples in the bottom row of \Fref{fig:dataset},
we present the consensus intersection over union (cIoU) metric 
{to take} multiple annotations into account.
Similar to the consensus metric in the VQA  task~\cite{kafle2017visual}, we assign scores to each pixel according to the consensus of multiple annotations. 

First, we convert the bounding box annotations to binary maps 
$\{\mathbf{b}_j\}_{j=1}^N$, where $N$ is the number of subjects. 
We extract a representative score map $\mathbf{g}$ by collapsing $\{\mathbf{b}_j\}$ across subjects but with considering consensus as
\begin{equation}
{\footnotesize
\mathbf{g} = \min \left(
\sum\nolimits_{j = 1}^N {\frac{{{\mathbf{b}_{j}}}}{\# \textrm{consensus}}} , 1\right)},
\label{eq:gtw}
\end{equation}
where $\#\textrm{consensus} \leq N$ is the minimum number of opinions to reach an agreement.
For each pixel in a score map $\mathbf{g}$, we compute the number of positive binary values (\ie, $\sum\nolimits_{j = 1}^N{\mathbf{b}_{j}}$).
If it is larger than or equal to \#\textrm{consensus}, then the pixel of $\mathbf{g}$ is set to a full score, \ie, 1.
Otherwise, it is set to a proportional score, which is less than 1.
Since we have three subjects, by majority rule, we set \#\textrm{consensus} to $2$ in our experiments.
Given this weighted score map $\mathbf{g}$ and predicted location response $\balpha$, we define the cIoU as
\begin{equation}
\small
\mathrm{cIoU}(\tau) = \frac{\sum\nolimits_{i \in \mathcal{A}(\tau)} {g_i}}{\sum\nolimits_i {g_i} + \sum\nolimits_{i \in \mathcal{A}(\tau) - \mathcal{G}} 1},
	\label{eq:IOUw}
\end{equation}
where $i$ indicates the pixel index of the map, $\tau$ denotes the threshold to determine positiveness of each pixel, $\mathcal{A}(\tau) = \{ i|{\alpha_i} {>} {\tau}\}$, and $\mathcal{G} = \{ i|g_i {>} 0\}$.
In \eqref{eq:IOUw}, $\mathcal{A}$ is the set of pixels with attention intensity higher than the threshold $\tau$, and $\mathcal{G}$ is the set of pixels classified as positives in weighted ground truth.
The denominator implies a weighted version of union of $\mathcal{A}(\tau)$ and $\mathcal{G}$.

\begin{figure}[t]
\centering

		\includegraphics[width=\linewidth]{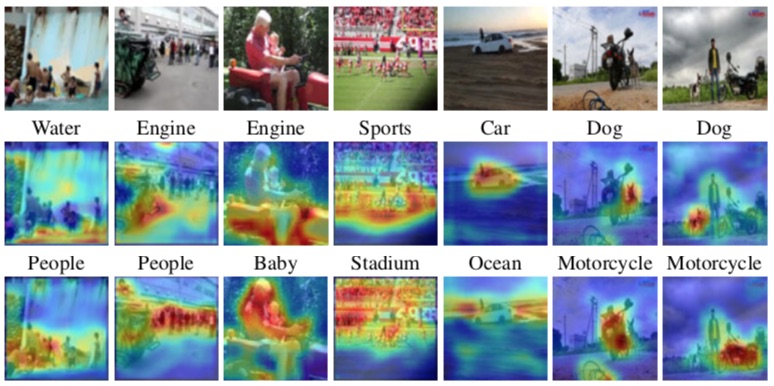}
\vspace{-6mm}
   \caption{\textbf{Interactive sound source localization.}
We show the responses of the network to different sounds while keeping the frame same. These results show that our network can localize the source of the given sound interactively. Label indicates the context of the sound. We do not use explicit labels. \vspace{-3mm}}
\label{fig:interesting}
\end{figure}

\begin{figure}
\centering

		\includegraphics[width=\linewidth]{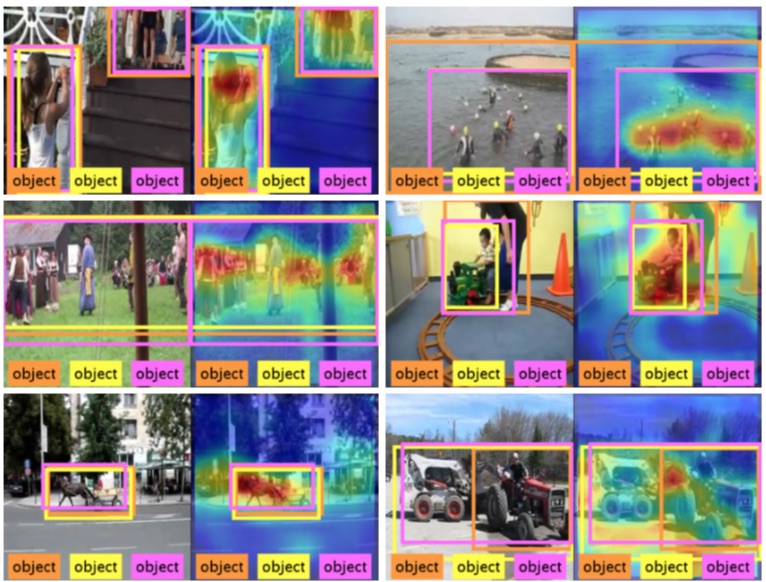}
\vspace{-6mm}
   \caption{\textbf{How well can our network localize the sound sources compare to humans?} Qualitative comparison of localization between our network and human annotations. {Human annotations (ground-truth) are represented by bounding boxes, and annotations from different subjects are indicated by different colors. The predictions from our method are the heat maps on the right panel of each block. We overlay human annotated bounding boxes on top of the heat maps for  comparisons.}
   }
\label{fig:loc_comparison}
\end{figure}

\paragrp{Qualitative analysis.}
We present the localization response 
$\balpha$ for qualitative analysis. 
\Fref{fig:heatmap} shows the localization results of the image-sound pairs from the Flickr-SoundNet dataset \cite{SoundNet} using the proposed \emph{unsupervised learning} approach. 
Our network learns to localize sound sources on a variety of categories without any supervision.
The sound sources are successfully localized in spite of clutters, and unrelated areas are isolated, \eg, in the ``water'' column of the \Fref{fig:heatmap}, people are isolated from the highlighted water areas. 
As shown in the ``concert'' examples; scenes include both stage people and the audiences. 
Even though they have similar appearances, the learned model is able to distinguish people on the stage from the audiences.

\begin{figure}[t]
\centering

		\includegraphics[width=\linewidth]{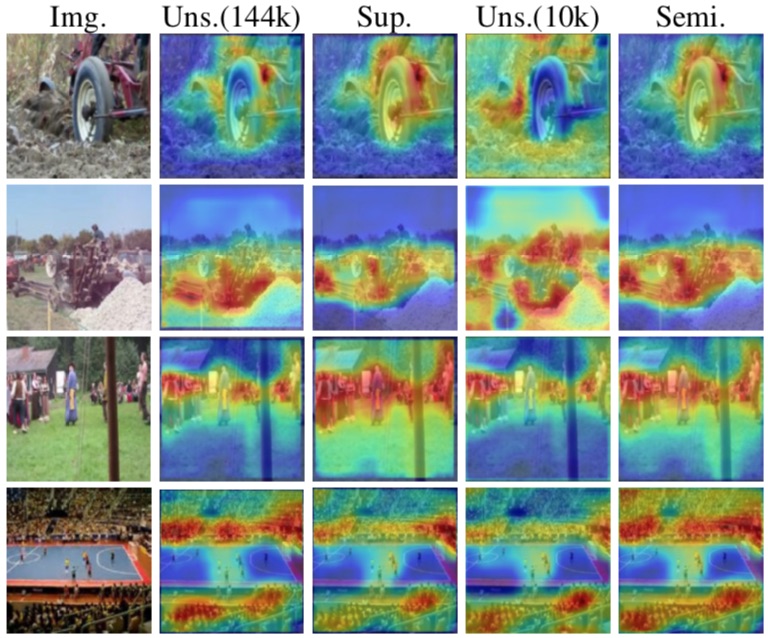}
\vspace{-6mm}
   \caption{\textbf{Qualitative sound localization results from different learning methods.}
We present the sound localization results from different learning methods. 
The supervised method generally localizes the sound source precisely due to the guidance of ground truths. 
Despite using less supervised data, the semi-supervised approach also gives comparably accurate localization results.}
\vspace{-3mm}
\label{fig:everything}
\end{figure}

At first glance, the results may look like hallucinating salient areas or detecting objects regardless of sound contexts.
It should be noted that our network responds interactively according to the given sound. 
\Fref{fig:interesting} shows examples of different input sounds for {the} same images where the localization responses change according to the given sound context. 
For a frame that contains water and people, when a water sound is given, the water area is highlighted. 
Similarly, the area containing people is highlighted when the sound source is from humans. 
With the network trained in the unsupervised manner, we qualitatively compare the localization performance with respect to human annotations. 
\Fref{fig:loc_comparison} shows sample qualitative results where the learned model performs consistently with human perception even though no prior knowledge is used.

While the network learns to localize sound sources in variety of categories without supervision, 
as aforementioned in \Fref{fig:pidgeon}, 
there are numerous cases that the unsupervised network falsely concludes the matching between visual and sound contexts. 
Using the semi-supervised scheme within the unified network model, we can transfer human knowledge in the form of supervision to remedy the pigeon superstition issue.
\Fref{fig:everything} shows the results by other learning methods. 
As expected, supervised learning methods 
localize objects more 
semantically aligned
with the 
ground truth supervision signals. 
We note that the proposed semi-supervised model achieves promising results by incorporating supervised and unsupervised data.

\begin{figure}
\centering

		\includegraphics[width=\linewidth]{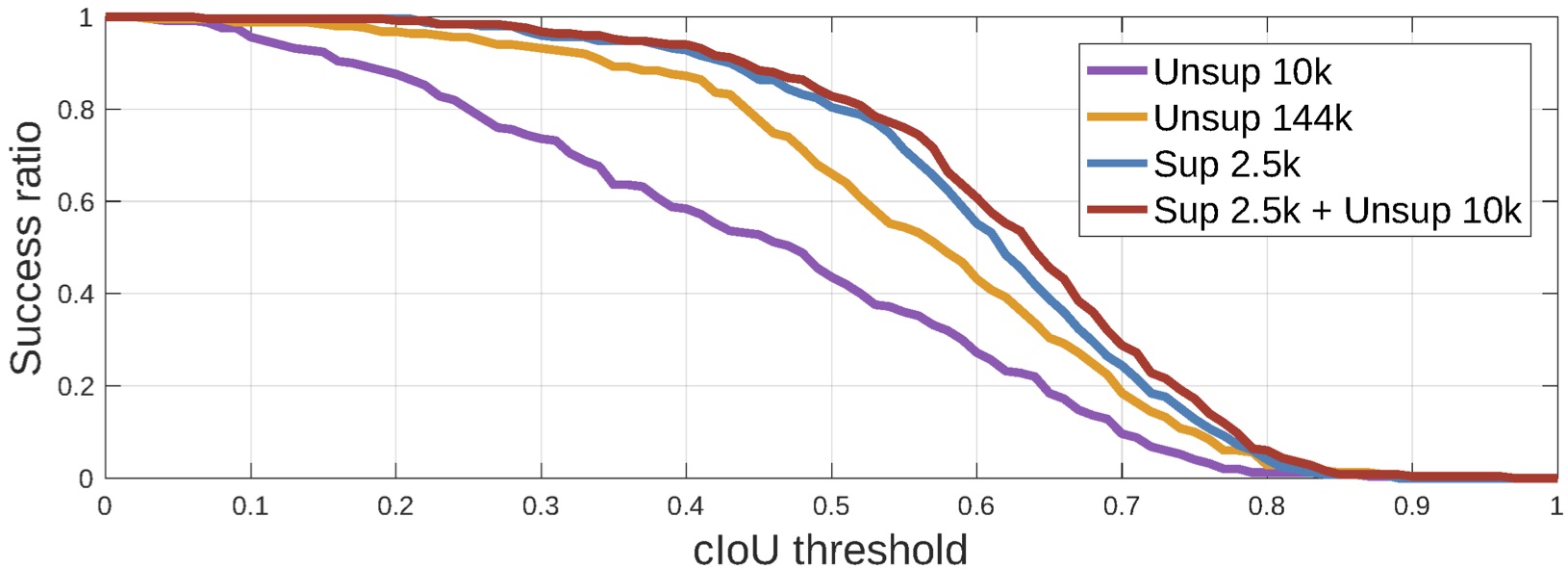}
        \vspace{-3mm}
		\caption{\textbf{Success ratio using varying cIoU threshold.} The attention mechanism with softmax without ReLU is used.}
\label{fig:supunsup}
\end{figure}
\paragrp{Quantitative results.}
\Tref{tab:supunsup} shows the success rates of cIoU and AUC for different learning schemes and {the} number of samples.
Using the common practice in object detection~\cite{everingham2010pascal}, we use $\tau=0.5$ for the cIoU threshold in (\ref{eq:IOUw}) to decide success or fail of the localization test.
We also report $95\%$ confidence intervals of the success rate of cIoU computed by binomial proportional confidence interval to see statistics.
{
We also present random prediction baseline results as reference.
Each experiment is repeated 100 times with random patterns and the statistics are computed. 
In addition, we compare with a method using a fixed center bounding box where the size is half of the image. 
Although this method achieves higher accuracy than the one with a random pattern, it still performs significantly worse than the proposed approaches.}
The results show that the unsupervised model with 10k samples learns meaningful knowledge from the sound and video pairs. 
Compared to our model trained in an unsupervised manner with 10k samples,
we observe significant improvement when the unsupervised network is trained with a larger number of samples, \ie, 144k samples.
We show the supervised learning results with 2.5k samples as reference.
Even the number of samples is smaller than the unsupervised method, the supervised model performs well.
When we train the network in the semi-supervised setting 
with both supervised and unsupervised losses, the model achieves the best performance.
The results suggest that there is complementary information from unlabeled data, which facilitates the model to generalize well.
We plot the success rate of the test samples in \Fref{fig:supunsup} according to cIoU thresholds.

\begin{table}[t]
	\caption{ \textbf{ Performance evaluation with different learning schemes.} 
	The cIoU measures the ratio of success samples at $\tau=0.5$ threshold. The AUC measures the area under the graph plotted by varying cIoU threshold from 0 to 1. }\vspace{1mm}\label{tab:supunsup}
	\centering
\resizebox{1\linewidth}{!}{
    \begin{tabular}{ lccccc }
		\toprule
		& \multicolumn{2}{c}{softmax} && \multicolumn{2}{c}{ReLU+softmax}\\
        \cmidrule{2-3} \cmidrule{5-6}
        & cIoU & AUC && cIoU & AUC\\
        \midrule
		Unsup. 10k		        & 43.6 $\pm$  6.2 & 44.9 &&  -- & --  \\
		Unsup. 144k	 	        & 66.0 $\pm$  5.7 & 55.8 &&  52.4 & 51.2 \\
		Sup. 2.5k			 	& 80.4 $\pm$  4.8 & 60.3 &&  82.0 & 60.7 \\
		Sup. 2.5k + Unsup. 10k	& 82.8 $\pm$  4.2 & 62.0 &&  84.0 & 61.9 \\
        \midrule
        \multirow{ 2}{*}{Baselines}  &  \multicolumn{2}{c}{cIoU} && \multicolumn{2}{c}{AUC} \\
        \cmidrule{2-3} \cmidrule{5-6} \\
        Random pattern ($\pm$ standard deviation) & \multicolumn{2}{c}{0.12 $\pm$ 0.2} && \multicolumn{2}{c}{32.3 $\pm$ 0.1} \\
        Random (Center attention - Half of the image size)	 	        & \multicolumn{2}{c}{23.2} && \multicolumn{2}{c}{40.7} \\
		\bottomrule
	\end{tabular}
    }
\end{table}

\begin{table}[t]
	\caption{ {\textbf{Performance comparison of learning methods with different amounts of data.}} }  \vspace{1mm}\label{tab:semisup}
	\centering
\resizebox{0.8\linewidth}{!}{%
    \begin{tabular}{ lccccc }
		\toprule
		& \multicolumn{2}{c}{softmax} && \multicolumn{2}{c}{ReLU+softmax}\\
        \cmidrule{2-3} \cmidrule{5-6}
        & cIoU & AUC && cIoU & AUC\\
        \midrule
        Unsup. 10k		& 43.6 & 44.9 &&  -- & --  \\
        Unsup. 144k	 	& 66.0 & 55.8 &&  52.4 & 51.2 \\
		\midrule
		Sup. 0.10k + Unsup. 10k  & -- & -- && 77.2 & 59.4 \\
		Sup. 0.25k + Unsup. 10k & -- & -- && 79.2 & 59.8  \\
		Sup. 0.50k + Unsup. 10k 	& 78.0 & 60.5 &&  79.2 & 60.3 \\
		Sup. 0.75k + Unsup. 10k & -- & -- && 80.4 & 60.5 \\
		Sup. 1.00k + Unsup. 10k	& 82.4 & 61.1 &&  82.4 & 61.1 \\
		Sup. 1.50k + Unsup. 10k 	& 82.0 & 61.3 &&  82.8 & 61.8 \\
		Sup. 2.00k + Unsup. 10k 	& 82.0 & 61.5 &&  82.4 & 61.4 \\
		Sup. 2.50k + Unsup. 10k 	& 82.8 & 62.0 &&  84.0 & 61.9 \\
		\midrule
		Sup. 2.50k + Unsup. 144k & -- & -- && 84.4 & 62.41 \\
		\bottomrule
	\end{tabular}
    }
    \vspace{-2mm}
\end{table}

\begin{table}[t]
	\centering
	\ra{1}
	\caption{ {\textbf{ Performance measure against individual subjects.}} }\vspace{1mm}
    	\label{tab:gt_ind}
    \resizebox{0.8\linewidth}{!}{%
	\begin{tabular}{lcccccccc}\toprule
		Subject & \multicolumn{2}{c}{Unsup. 144k} &  & \multicolumn{2}{c}{Sup.} &  & \multicolumn{2}{c}{Semi-sup.}\\
		\cmidrule{2-3} \cmidrule{5-6} \cmidrule{8-9}
		& IoU & AUC && IoU & AUC  && IoU & AUC \\\midrule
		Subj. 1 & 58.4 & 52.2 &&  70.8  &  55.6  &&  74.8 &  57.1  \\
		Subj. 2 & 58.4 & 52.4 &&  72.0  &  55.6  &&  73.6 &  57.2  \\
		Subj. 3 & 63.6 & 52.6 &&  74.8  &  55.6  &&  77.2 &  57.3  \\
		Avg. 	& 60.1 & 52.4 &&  72.5  &  55.6  &&  75.2 &  57.2  \\
		\bottomrule
	\end{tabular}
    }
    \vspace{-2mm}
\end{table}

We analyze the effect of the number of labeled samples using the semi-supervised setting.
\Tref{tab:semisup} shows that near 1k supervised samples are sufficient for the semi-supervised model to learn well.
We note that the proposed model benefits more from a combination of both types of data than simply increasing the number of supervised samples.
{Furthermore, increasing the number of unsupervised samples in {the} semi-supervised setting, \ie, Sup. 2.5k + Unsup. 144k samples, shows marginal improvement. 
}

\begin{table}[t]
	\centering
	\ra{1}
	\caption{ {\textbf{ Performance measure using different test sets.}} }\vspace{1mm}
    	\label{tab:crossval}
    \resizebox{1.0\linewidth}{!}{%
	\begin{tabular}{lcccccccc}\toprule
		Test set & \multicolumn{2}{c}{Set1} &  & \multicolumn{2}{c}{Set2} &  & \multicolumn{2}{c}{Set3}\\
		\cmidrule{2-3} \cmidrule{5-6} \cmidrule{8-9}
							   & cIoU & AUC  &&  cIoU  &  AUC   &&  cIoU &  AUC   \\ \midrule
		Sup. 0.5k + Unsup. 10k & 78.0 & 60.5 &&  73.6  &  60.0  &&  76.4 &  61.7  \\
		Sup. 1.0k + Unsup. 10k & 82.4 & 61.1 &&  77.2  &  60.8  &&  77.9 &  62.1  \\
		Sup. 1.5k + Unsup. 10k & 82.0 & 61.3 &&  79.7  &  62.2  &&  76.8 &  62.3  \\
		Sup. 2.0k + Unsup. 10k & 82.0 & 61.5 &&  78.3  &  61.9  &&  78.6 &  62.5  \\
        Sup. 2.5k + Unsup. 10k & 82.8 & 62.0 &&  80.8  &  62.2  &&  79.0 &  62.5  \\
		\bottomrule
	\end{tabular}
    }
    \vspace{-2mm}
\end{table}

To analyze the subjectiveness of supervision,
we report the IoU performance of each annotator independently in \Tref{tab:gt_ind}.
While the numbers across subjects vary slightly, the variance is small and the performance trends are consistent among the methods.
This suggests that cIoU is an effective measure.
Furthermore, despite the ambiguity nature of the localization task, our method performs coherently with the human perception in images.

We show cross validation results using two more splits (cross validation sets) in \Tref{tab:crossval}. 
The Set1 is the test set 
used in our early work \cite{senocak2018learning} and in the previous experiments. 
The other two sets are selected randomly from our annotated dataset for additional evaluation. 
The sets introduced here are mutually exclusive and each set has the same number of samples. 
We present performance for cross-validation sets with the same semi-supervised network approach. 
For each set evaluation, a network model is trained from scratch using the dataset excluding the test samples. 
These results {show} consistent trends, except {for} the slight variation on the number of necessary supervised samples for performance improvement.
{We also conduct an ablation study to analyze the weight between the supervised and unsupervised losses.
\Tref{table:weight_v2} shows the accuracy for the Unsup. 144k $+$ Sup. 0.5k setting. 
The proposed method with the balance weights of {$\{\textrm{Unsup.}=0.5, \textrm{Sup.}=0.5\}$ and $\{\textrm{Unsup.}=0.75, \textrm{Sup.}=0.25\}$} with a large number of {data} performs best.}

\begin{table}[!h]
\caption{\textbf{Performance measure against different loss balance weights with the ``Unsup. 144k + Sup. 0.5k''  data setup.}
}
\centering
\resizebox{0.55\linewidth}{!}{%
\begin{tabular}{c c c}
\toprule
Loss weights & cIoU & AUC \\ [0.5ex]
\midrule
Unsup.=0.1, Sup.=0.9     & 82.8                       & 62.1                     \\ 
Unsup.=0.25, Sup.=0.75     & 84.0                       & 62.0                     \\
Unsup.=0.5, Sup.=0.5     & 85.2                       & 62.36                     \\
Unsup.=0.75, Sup.=0.25     & 85.2                       & 62.64                     \\
Unsup.=0.9, Sup.=0.1     & 83.2                       & 62.3                     \\
\bottomrule
\end{tabular}
}
\label{table:weight_v2}
\end{table}
\subsection{Ambient Sound and Learned Embeddings}
\begin{figure}[!t]
\vspace{-1mm}
\centering

		\includegraphics[width=\linewidth]{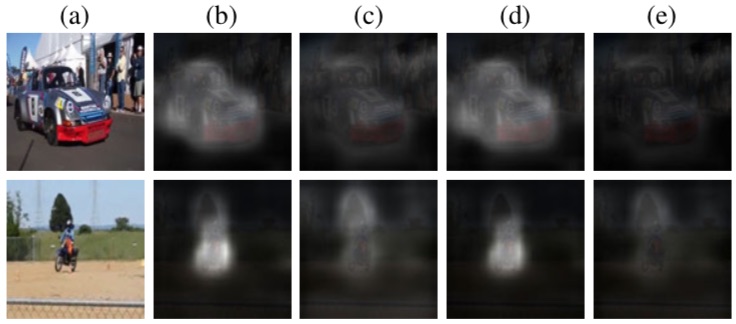}
\vspace{-4mm}
\caption{\textbf{Ambient sound results.} We show examples of frames with ambient sounds. 
(a) sampled input frames.
(b) location responses against object indicating sound in \emph{Softmax} only attention mechanism.
(c) location responses against ambient sound in \emph{Softmax} only attention mechanism.
(d) location responses against object indicating sound in \emph{ReLU+Softmax} attention mechanism.
(e) location responses against ambient sound in \emph{ReLU+Softmax} attention mechanism.
The proposed network ouputs discernible confidences between object-like and ambient sounds.}
\label{fig:ambient}
\end{figure}

\paragrp{Ambient sound analysis.}
We analyze the proposed method with non-object and ambient sounds (\eg, environmental sounds, wind sounds, background activities, and narration). 
We feed the frames with one of these ambient sounds into our network to see how it reacts.
\Fref{fig:ambient} shows that the proposed method gives noticeably low confidence scores to ambient sound, and high reaction to the object indicating sound.
\Fref{fig:ambient} shows that the method based on 
ReLU+softmax performs better on ambient sounds. 
This is due to the ReLU operation that clips the negative values in an attention map to zero in the training phase.
Our attention map is computed based on inner products between normalized vectors of which range is in $[-1,1]$.
For the method with  ReLU+softmax, 
the negative values are clipped to 0.
Consequently, the method with ReLU+softmax suppresses uncorrelated sound responses well. 
We show the attention response before softmax to show absolute (\ie, non-relative) values.
The responses of ambient sound {are} relatively weaker than object sounds. 
We use gray scale heatmaps in \Fref{fig:ambient} for better illustration.
While this is out of the scope of this work due to the requirement of human annotations, the proposed model helps {to deal} with off-context sound cases.

\paragrp{Learned embeddings.}
Our network generates embeddings which 
can be used to analyze the effectiveness of learned representations.
As discussed in~\Sref{sec:learning}, our network is trained to have similar predictions from sounds and images by mapping to close locations in the learned embedding space when both modalities have similar semantic contents.
For example; if properly learned, the embeddings of soccer match images will be close to those of other sports games, but not to the embeddings of the instrument lessons. 
For ease of illustration, we slightly abuse the notations of the visual and sound embeddings as a functional form, \ie, $\mathbf{f}_v(X_v, X_s)$ and $\mathbf{f}_s(X_s)$, where $X_{\{v, s\}}$ denote a video frame and a sound waveform of an input sample $X$, respectively.
We note that these two embeddings are encouraged to have a shared space that allows them to be comparable by metric learning.
Thus, we can directly compare heterogeneous embeddings. %
We carry out all the experiments on the Set1 subset in~\Tref{tab:crossval}, and denote it as database $\mathcal{D}$.
\begin{table}
\caption{\textbf{Evaluation of cross-modal k-nearest neighbor search with pseudo labels.} The success ratios are calculated for each sample and average scores of each case are reported.
}
\centering
\resizebox{0.6\linewidth}{!}{%
\begin{tabular}{c c c}
\toprule
Top-k & Image $\rightarrow$ Audio &Audio $\rightarrow$ Image \\ [0.5ex]
\midrule
Top-20     & 77.8                       & 66.6                     \\ 
Top-15     & 79.1                       & 67.7                     \\ 
Top-10     & 80.8                       & 69.9                     \\ 
\midrule
Random-10     & 38.2                       & 38.1                     \\[0.5ex]
\bottomrule
\end{tabular}
}
\label{table:ret1}
\end{table}
We analyze the semantic quality of the embeddings in \Tref{table:ret1}, where the sound query based video retrieval and vice versa are conducted and we report the success ratio of semantically meaningful {matches}.
Given the query $X$, we conduct the $k$-nearest neighbor search by measuring the distance $d(\mathbf{f}_s(X_s), \mathbf{f}_v(Y_v, Y_s))$ over samples $Y$ in the database $\mathcal{D}$, \ie, $Y\in \mathcal{D}$, where $d(\cdot)$ denotes the cosine distance in that we empirically found the performance is higher with it. 
However, since we do not have ground truth information, instead we use a pseudo label approach by obtaining the top-10 label predictions of each sample from the pretrained VGG-16~\cite{Simonyan15} and SoundNet~\cite{SoundNet} according to the modality type, and use them as pseudo labels.
We %
{consider} the %
{successful} match when 
the intersection set of the pseudo labels between the query and the k-nearest neighbors is not empty, \ie, if they have at least one shared prediction label, and the %
{failure} otherwise. 
{We compute the random chance based on random trials (ideal random chance cannot be obtained due to unknown true classes). 
For each sample, we randomly select 10 samples from our database for experiments. 
We repeat this experiment 100 times and report the average score. 
The same procedure is carried out for both cross-modal directions, \ie, Image $\rightarrow$ Audio and Audio $\rightarrow$ Image.
The reason for the performance top-20$<$top-15$<$top-10, is that, in the limited retrieval set, there are classes 
of which the number of samples is less than $n$. 
If $n\ll k$, more samples which have unrelated content appear in the {top} $k$ number of retrieved samples. 
}

\begin{figure}[!t]
\centering

		\includegraphics[width=\linewidth]{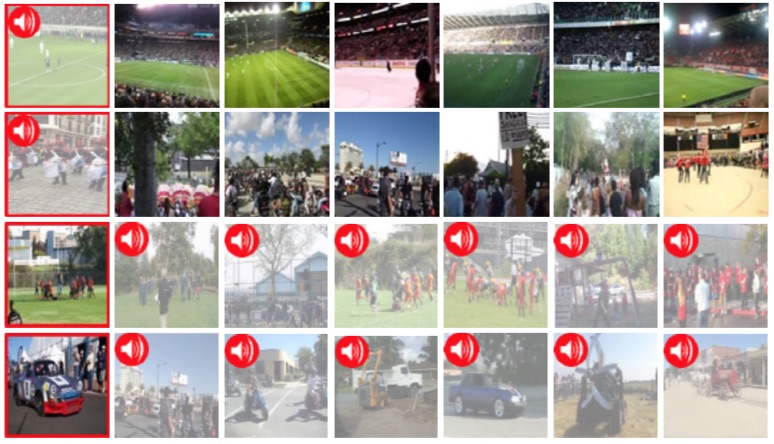}
\vspace{-4mm}
   \caption{{\textbf{Semantically relevant neighborhood of the given sample in cross-domain.}} Each row shows one query and k-nearest neighbors. Red color borderline indicates the query sample and sound icon indicates the sound modality, where whiten images indicate no visual information is used but are overlaid for reference.
   Nearest neighbors to the query in the shared embedded space are the ones which have the most similar contextual information to the query.}  
\vspace{-3mm}
\label{fig:retrieval}
\end{figure}
\begin{figure}[!t]
\centering

		\includegraphics[width=\linewidth]{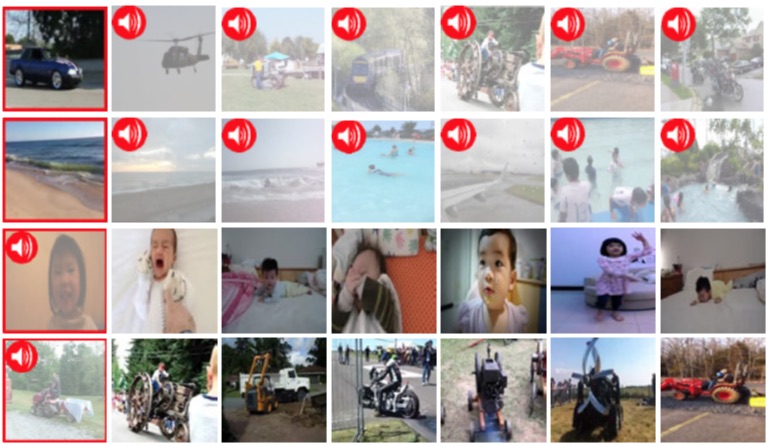}
\vspace{-4mm}
   \caption{{\textbf{Associative  behavior  of  the embeddings with the conditional input.}} Given query $X_{v,s}$ and selected modality of it as in {the} first column, $\mathbf{f}_v$ for each sample in the dataset is computed by conditioning on keeping the selected modality information same as query but using the cross-modality information of the sample as a corresponding pair; $d(\mathbf{f}_v(X_v, X_s), \mathbf{f}_v(X_v, Y_s))$ (the top half of the figure, where visual modality is not used for gallery samples, \ie, indicated by whiten images) or $d(\mathbf{f}_v(X_v, X_s), \mathbf{f}_v(Y_v, X_s))$ (the bottom half of the figure, where sound modality is not used for gallery samples). Nearest samples have the most similar semantic context in cross-domain.
   }
\vspace{-3mm}
\label{fig:another_retrieval}
\vspace{-2mm}
\end{figure}

\Fref{fig:retrieval} shows 
the neighboring samples that match with queries semantically. 
In the third row, our model not only locates the ``football'' samples close to each other, but also maps the scenes with ``a group of people'' or  ``a group of people on the green field'', where the query can also be perceptually seen as a group of people on the grass.

In addition, we analyze the associative behavior of the embeddings $\mathbf{f}_v$ according to the different input in \Fref{fig:another_retrieval}.
Specifically, given the query $X$, we conduct the same experiment as above, but by using  $d(\mathbf{f}_v(X_v, X_s), \mathbf{f}_v(X_v, Y_s))$ or  $d(\mathbf{f}_v(X_v, X_s), \mathbf{f}_v(Y_v, X_s))$.
For the first case, 
since every frame is same but the corresponding sounds are different according to samples in the database, closest neighbors to the query are the ones {that} have similar audio information to the query. 
Note that this is different from  sound retrieval because, by the association $\mathbf{f}_v(X_v, Y_s)$, we expect that context information in $Y_s$ irrelevant to $X_v$ is discarded. 
The second case is by keeping the sounds same but using different frames.
In this scenario, we expect that visual context information in $Y_v$ irrelevant to the sound context in $X_s$ is discarded, so that the selected semantic context {is} retrieved.
The results show that our model performs well in
sound localization with 
conditional input, and learns semantic audio-visual correspondence.
%
\section{Video Applications}
\label{sec:videos}

\begin{figure}[!b]
\centering

		\includegraphics[width=\linewidth]{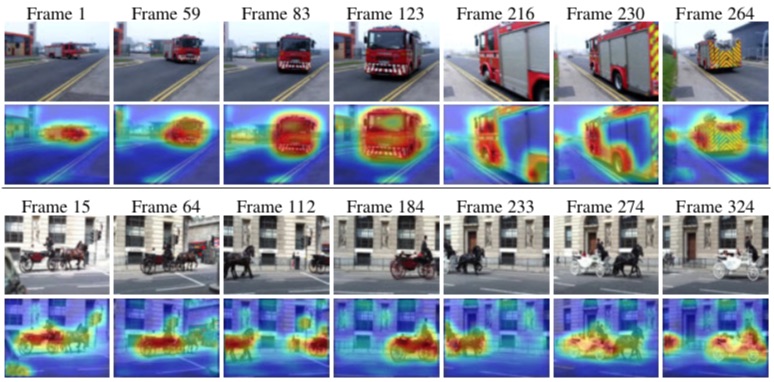}
\vspace{-4mm}
   \caption{{\textbf{ Sound source localization in video sequences.} We show the results of localizing sound sources in the video sequences. No temporal information is used. Each frame is processed independently.}}
\label{fig:tracking}
\end{figure}

We show the localization results not only on still images but on videos as well. 
Each video frame and {the} corresponding audio, which is obtained from a sliding window, are processed  independently without using motion or temporal information (although adding temporal cues can further improve  localization results~\cite{owens2018audioECCV}).
\Fref{fig:tracking} shows the proposed method highlights the sound sources despite fast motion, cluttered and complex scenes, changes of appearance as well as size, even without utilizing a temporal cue. 
We also apply this proposed model to sound based 
360$\degree$~video camera view panning. Details are explained in the next section.
\begin{figure*}
\centering

		\includegraphics[width=\linewidth]{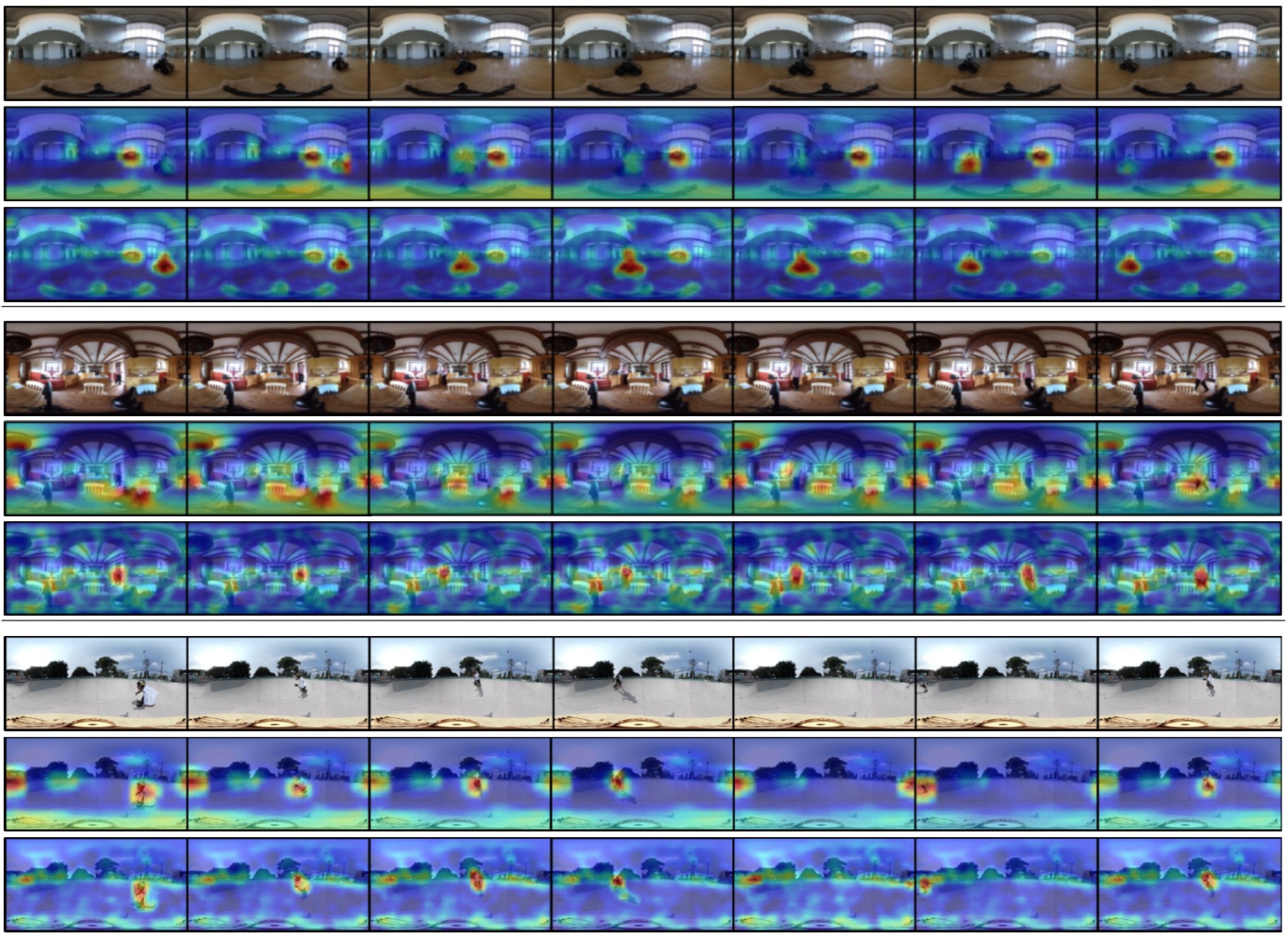}
\vspace{-4mm}
   \caption{{\textbf{Qualitative comparison of visual and audio based saliency maps.}} Consecutive frames of videos are shown in the first row. Vision based saliency maps computed from \cite{Cheng2018cubemapsaliency} are presented in the second raw. Our audio based saliency results are in the third row. It shows sound itself carries out rich information in 360\degree~videos{,} and the proposed method performs as well as the vision based method to predict saliency maps.}
\label{fig:360_comparison}
\end{figure*}

\begin{figure*}
\centering

		\includegraphics[width=\linewidth]{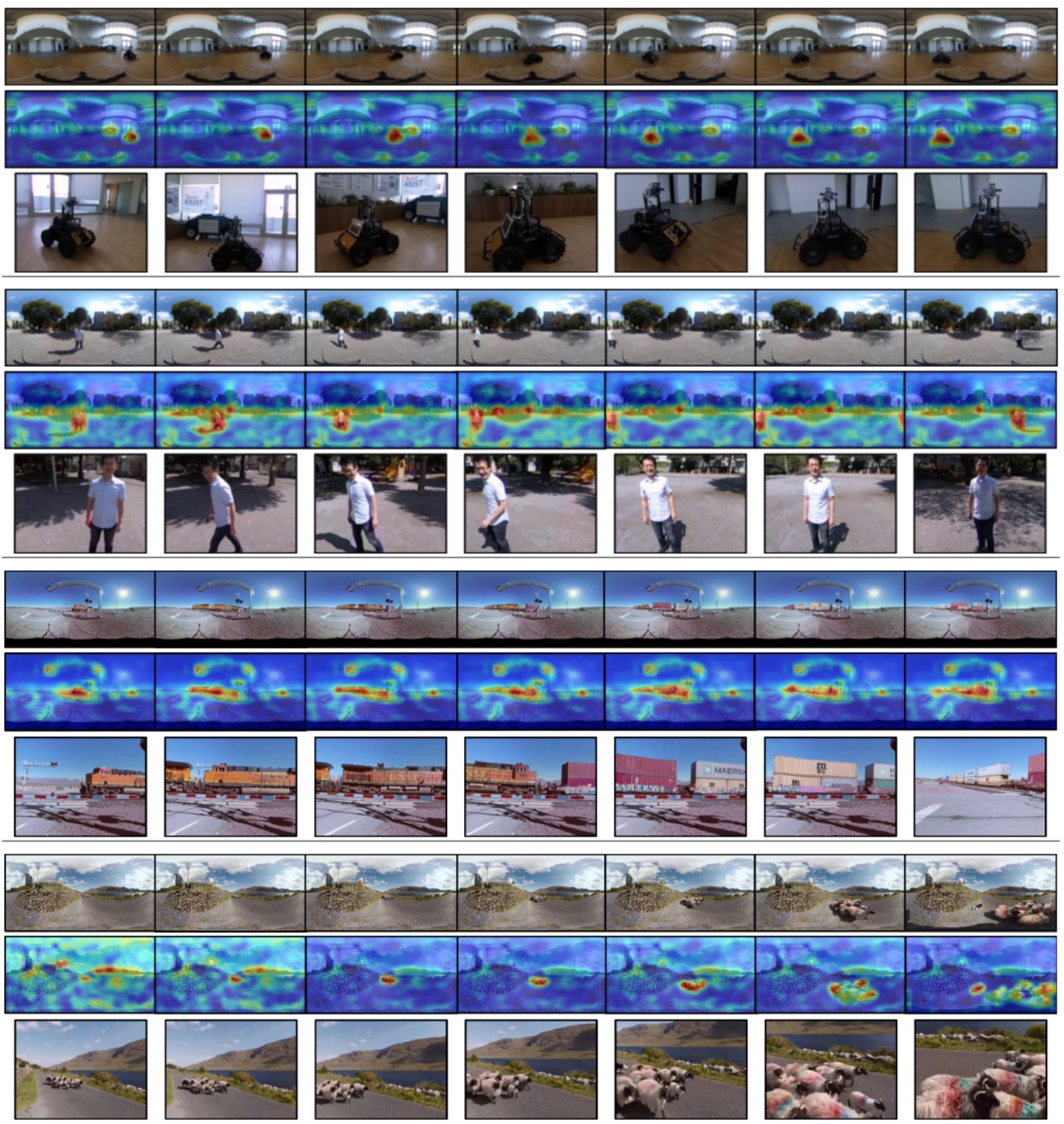}
\vspace{-2mm}
   \caption{{\textbf{Normal Field-of-View (NFoV) trajectory results.}} We present consecutive video frames and corresponding audio based saliency maps in the first and second rows respectively. NFoVs in multiple time steps that are used for 360\degree~content exploration are shown in the last row. Results show that audio based saliency maps can be effectively used for camera view panning in 360\degree~videos.}
\label{fig:360_NFoV}
\end{figure*}

\paragrp{Automatic camera view panning in 360\degree~videos.}
Recently, 360\degree~videos have become emerging media as rich immersive contents are captured. 
These videos cover {a} wider field of view than typical perspective cameras, and are supported by video streaming websites.
With 360\degree~videos, users can easily view the contents in any direction by navigating projective normal Field-of-View (NFoV).
Nevertheless, it is often cumbersome to figure out where and what to watch in the videos by choosing the viewing angles manually. 
This process requires manual and  exhaustive exploration of the space to find events during the full duration of a  video. 

Several methods have been recently developed for
navigating 360\degree~videos by finding NFoV of interest.
These methods mainly leverage visual information such as saliency \cite{su2016pano2vid, su2017CVPRpano2vid, Cheng2018cubemapsaliency, HuLinCVPR17deepPilot, Chou17grounding360}. 
However, we tackle this problem from the perspective of audio sources. 
We observe that visual events are usually accompanied {by} sounds, and humans {use not only} visual information but also audio cues to rapidly focus on sights of interest~\cite{jones1975eye}.
Thus, as visual cues are important for full scene awareness in 360\degree~videos, sound cues are also crucial. 
For 360\degree~videos,
we leverage those accompanied sounds 
to guide the navigation direction.
To the best of our knowledge, this is the first attempt to use sound cues for the automatic 360\degree~video navigation.

We extract the frames at 30 frames per second (FPS) in an equirectangular image format, of which the image resolution is $480\,{\times}\,960$. 
We feed the equirectangular images directly to our network which is pretrained on the Flickr-SoundNet dataset under the semi-supervised setting. 
We use the same procedure described in Sections~\ref{sec:soundnet} and \ref{sec:vis_model}. 
Since the pretrained network is fully convolutional, we can apply it to the original resolution as it is without changing the aspect ratio of the input frame.
For each time stamp, we feed a sliding window of sound and a frame associated with the center of the window. 
We use the identical procedure in \Sref{sec:att_model} to obtain the frame based sound saliency maps, \ie, sound localization response maps.
\Fref{fig:360_comparison} shows qualitative results of our sound guided saliency maps with comparison to the recent method  \cite{Cheng2018cubemapsaliency} which uses the visual cue alone for the 360\degree~video exploration task. 
{We conduct a user study, where each subject is asked to select the preferred one from two saliency videos (presented in random shuffle) generated by the method based only on vision \cite{Cheng2018cubemapsaliency} and our algorithm. 
{We design an interface for users to watch and hear the original video and to select 
the best one based on the following criteria}:
1) The one activates in most of the objects/areas corresponding to the dominant sound in audio, \ie, dominant sound source in the scene.
2) The one activates in regions corresponding to the ``dominant'' content of audio more accurately,
3) The one best localizes the sound source according to the above criteria and longer duration in the entire video.
4) If there is no perceived difference between given saliency maps, then the user is expected to pick the $[Similar]$ option.
We obtain the relative scores between every pair of evaluated methods. We collect the results of 11 videos from 30 participants.   
The results are presented in~\Tref{table:user_study}.}
\begin{table*}
\caption{\textbf{User study of 360$\degree$~video saliency maps.} 
The voting rates (\%) are reported for each video used in the user study.
}
\centering
\resizebox{0.8\linewidth}{!}{%
\begin{tabular}{c  c c c c c c c c c c c  c}
\toprule 
{Video title} & People & Sheep & Ocean Beach & Drumming & Train & Robot1 & Skate & Orange Helicopter & Kitchen Man & Red Car & VR Security & Average\\ [0.5ex]
\midrule
{Ours} & \textbf{96.7}& \textbf{43.3}& \textbf{66.7}& \textbf{86.7}& \textbf{93.3}& \textbf{73.3}& 26.7         & \textbf{53.3}&\textbf{83.3}& \textbf{86.7}& \textbf{96.7}& \textbf{73.3}\\
Cheng~\etal\cite{Cheng2018cubemapsaliency} & 3.3 & 30.0& 26.6& 10.0& 6.7 & 13.4& 26.7& 36.7& 10.0& 3.3 & 0.0 & 15.2 \\
{Similar} & 0.0 & 26.7         & 6.7          & 3.3          & 0.0          & 13.3         & \textbf{46.6}& 10.0         & 6.7         & 10.0     & 3.3  &11.5\\
\bottomrule
\end{tabular}
}
\label{table:user_study}
\end{table*}

We note that the experimental comparisons here are only to show how different modalities respond differently to the same input content. 
The results show that sound can carry rich information in 360\degree~contents so that the proposed audio-visual method performs well as much as the sophisticated vision only method and even better in some scenarios that the vision based method cannot perform well, such as the second example in~\Fref{fig:360_comparison} where an old man is walking around in the kitchen and talking at the same time. {The} vision based method focuses on the objects in the kitchen because it uses objectness information to predict saliency. However, {the} proposed audio-visual method can capture the speaking man.  
While we directly use our pretrained network, which has not been trained on the equirectangular images, 
it works plausibly well without any additional fine-tuning.
The quality would be further improved by using the cube map coordinates~\cite{Kopf2016,Cheng2018cubemapsaliency}, but the simple equirectangular format was sufficient for our examples.

%

A obtaining the saliency maps, we generate NFoV trajectories based on the selected interesting areas as shown in \Fref{fig:360_NFoV}.
We use the AUTOCAM~\cite{su2016pano2vid} method to generate a path of the sound source in 360\degree~videos.
We use the weighted average of pixel locations from our saliency maps as an importance measure for each frame to estimate the center of the region of interest, instead of selecting regions based on a binary map of visual importance as done in \cite{su2016pano2vid,su2017CVPRpano2vid}.
We apply {this approach} to videos that contain different types of sound sources {such as moving or stationary, slow or fast moving, as well as small or large objects.} 
More results can be found in the supplementary materials.
%
%
\section{Discussion and Conclusion}\vspace{-1mm}
We tackle a new problem, learning based sound source localization in visual scenes, and construct its new benchmark dataset.
By empirically demonstrating the capability of our unsupervised network, 
we show the model plausibly works in a variety of categories but partially, in that, without prior knowledge, the network can often get to {a} false conclusion.
We also show that leveraging {a} small amount of human knowledge can discipline the model, so that it can correct to capture semantically meaningful relationships.
These may imply that, by the definition of learnability~\cite{shalev2014understanding}, the task is not a fully learnable problem only with unsupervised data 
{in our setting, which is static-image based single-channel audio source localization}, but can be fixed with even a small amount of relevant prior knowledge. 
{Although the sound localization task is not effectively addressed with our unsupervised learning approach with static images and mono audios, other methods that use spatial microphones~\cite{gao2019visual-sound,acoustic_camera2019,morgadoNIPS18,Zunino} or temporal information, motion~\cite{owens2018audioECCV} and synchronization~\cite{korbar2018Cooperative}, with multiple frames have been shown to perform well on this task with unsupervised algorithms.}
In the following, we conclude our work with additional discussion for future investigation.

\paragrp{Representation learning}
The results and the conclusion made in this work may allow us to deduce the way of machine understanding about sound source localization in visual scenes.
For example, in unsupervised representation learning from sound-video pairs~\cite{Ambient,Arandjelovic17}, 
our results may indicate that
some of the representations behave like the pigeons (as in the second row of the ``Railway'' column in Figure 5 of Arandjelovic~\etal\cite{Arandjelovic17}), and suggest that at least a small amount of supervision should be incorporated for proper sound based representation learning.
Additionally, this work would open many potential directions for future research, \ie, 
multi-modal retrieval, sound based saliency, representation learning and its applications.

\paragrp{Noisy unsupervised dataset}
We use an ``in-the-wild'' dataset,  Flickr-SoundNet~\cite{SoundNet}, which contains noise and outliers. 
As with many other self-supervised and unsupervised methods \cite{arandjelovic2018objects,owens2018audioECCV}, our method also does not explicitly handle such outlier effects.
Despite the fact, our method works plausibly.
Although training neural networks robust to noise and outlier data is still an open problem, the performance and the quality of learned representation could be further improved by adopting robust mechanisms.

\section*{Acknowledgment}
A. Senocak, J. Kim and I.S. Kweon were supported by the National Information Society Agency for construction of training data for artificial intelligence (2100-2131-305-107-19).
M.-H. Yang is supported in part by
NSF CAREER (No. 1149783).
T.-H. Oh and I.S. Kweon are the co-corresponding authors of this work.

%

%


%

%
%
%

\ifCLASSOPTIONcaptionsoff
\newpage
\fi

{\small
	\bibliographystyle{IEEEtran}
	\bibliography{egbib}
}
\newpage
\renewcommand\thesection{\Alph{section}}
\newpage
\noindent {}
\newpage
\noindent {\LARGE \textbf{Appendix}}
\section*{Supplementary Material}
This document contains material complementary to the manuscript, mainly on the qualitative results of our {cross-domain} k-nearest neighborhood search for learned embeddings and 360\degree~video applications. More results can be found in the supplementary video (available at \url{https://youtu.be/gDW8Ao8hdEU}). Figures with higher resolution are available at \url{https://drive.google.com/open?id=1HuyJmvYvxrEkgMnAQE9KjfP6hwOovJvq}.

\setcounter{section}{0}
\section{{Cross-Domain} k-Nearest Neighborhood Search on Learned Embedding Features}
\begin{figure*}[!t]
\vspace{-10mm}
\includegraphics[width=\linewidth]{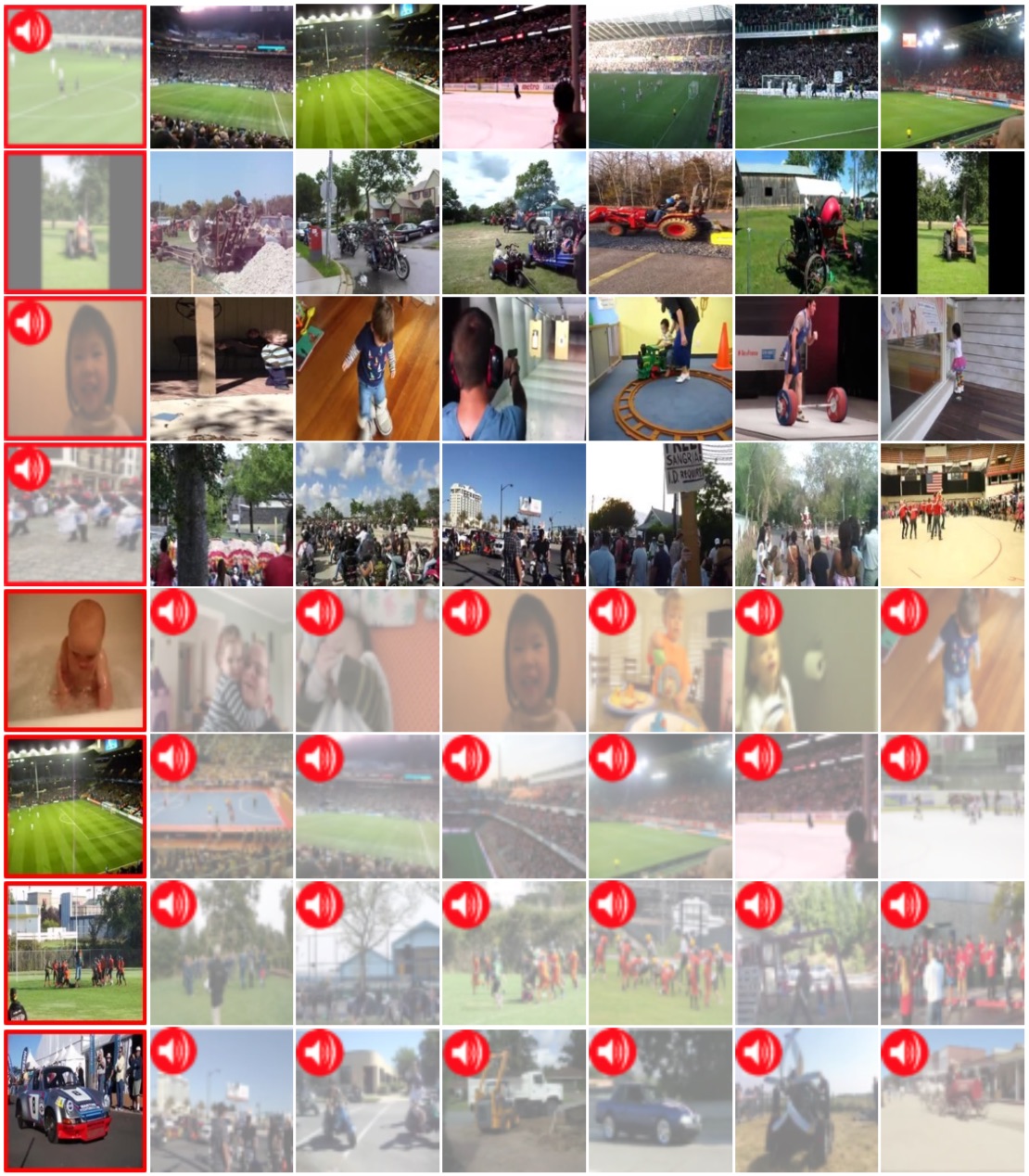}
\vspace{-2mm}
   \caption{{\textbf{Semantically relevant neighborhood of the given sample in cross-domain.}} Each row shows one query and k-nearest neighbors. Red color borderline indicates the query sample and sound icon indicates the sound modality, where whiten images indicate no visual information is used but are overlaid for reference.
   Nearest neighbors to the query in the shared embedded space are the ones which have the most similar contextual information to the query.}
\vspace{-3mm}
\label{fig:retrieval_supp}
\end{figure*}

\begin{figure*}
\vspace{-10mm}
\includegraphics[width=\linewidth]{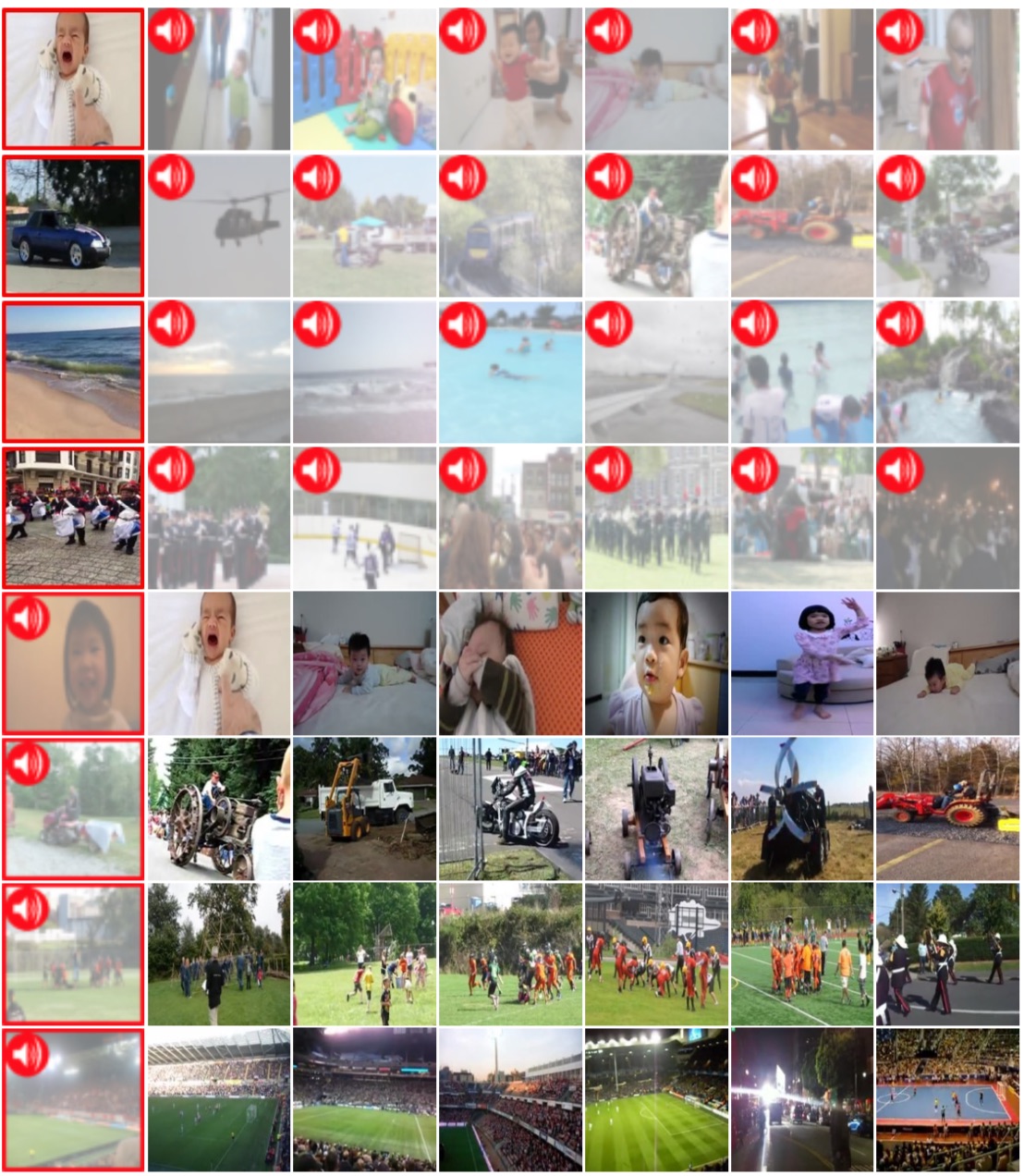}
\vspace{-2mm}
   \caption{{\textbf{Associative  behavior  of  the embeddings with the conditional input.}} Given query $X_{v,s}$ and selected modality of it as in {the} first column, $\mathbf{f}_v$ for each sample in the dataset is computed by conditioning on keeping the selected modality information same as query but using the cross-modality information of the sample as a corresponding pair; $d(\mathbf{f}_v(X_v, X_s), \mathbf{f}_v(X_v, Y_s))$ (the top half of the figure, where visual modality is not used for gallery samples, \ie, indicated by whiten images) or $d(\mathbf{f}_v(X_v, X_s), \mathbf{f}_v(Y_v, X_s))$ (the bottom half of the figure, where sound modality is not used for gallery samples). Nearest samples have the most similar semantic context in cross-domain.}
\vspace{-3mm}
\label{fig:another_retrieval_supp}
\end{figure*}

We show cross-modal neighborhood search on our embedded %
features for different queries. 
Figures~\ref{fig:retrieval_supp} and~\ref{fig:another_retrieval_supp} show sample qualitative results. 
Each figure shows the results of each approach that we introduce in Section 5.~Our model generates aligned features which help to project semantically related cross-modal samples into same neighborhood in the shared space.

\section{{Audio based} Saliency Prediction in 360\degree~Videos and Comparison with Vision based Saliency}
\begin{figure*}[t]
\centering
\includegraphics[width=0.9\linewidth]{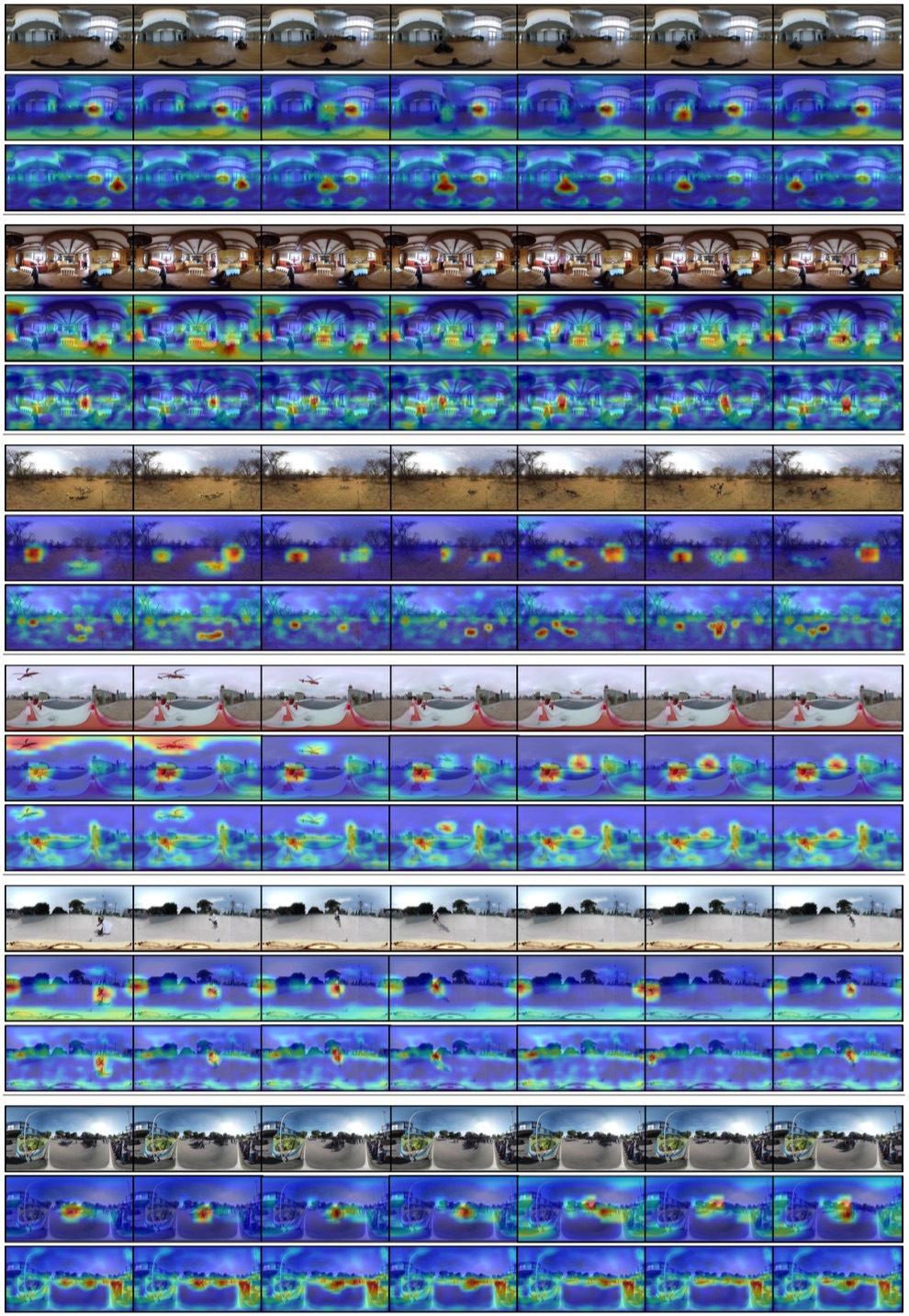}
\vspace{-2mm}
   \caption{{\textbf{ Qualitative comparison of visual and audio based saliency maps.}} Consecutive frames of videos are shown in the first row. Vision based saliency maps computed from \cite{Cheng2018cubemapsaliency} are presented in the second raw. Our audio based saliency results are in the third row. It shows that sound itself carries out rich information in 360\degree~videos{,} and the proposed method performs as well as vision based method to predict saliency maps.}
\label{fig:360_comparison_supp}
\end{figure*}

As discussed in Section 6 
of the manuscript, here
we compare our {sound based} saliency prediction results with vision based method~\cite{Cheng2018cubemapsaliency} qualitatively to show that sound gives informative saliency maps as vision based saliency methods. 
These results are illustrated in~\Fref{fig:360_comparison_supp}.

\section{360\degree~Videos Navigation}
\begin{figure*}[!t]
\includegraphics[width=0.95\linewidth]{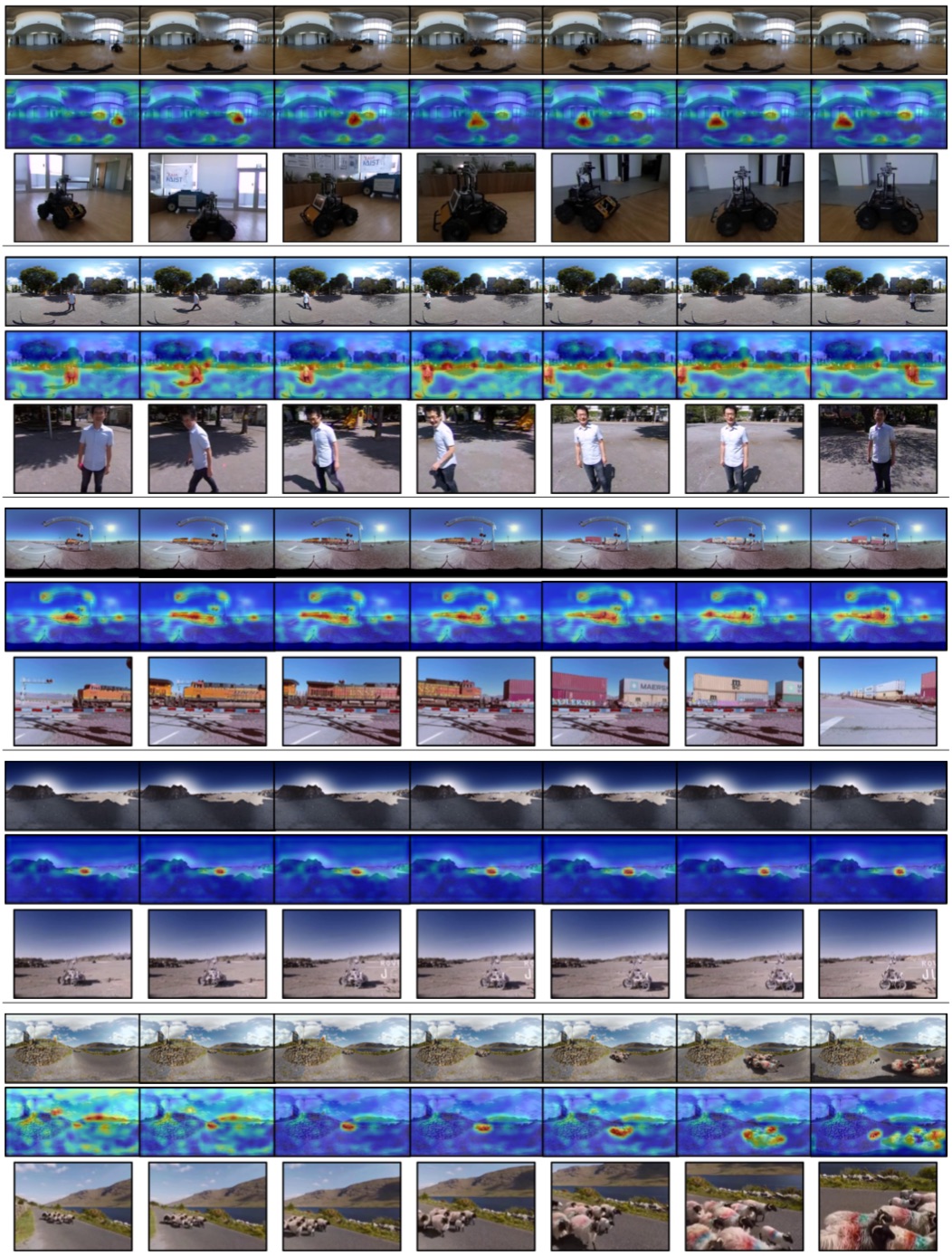}
\vspace{-2mm}
   \caption{{\textbf{Normal Field-of-View (NFoV) trajectory results.}} We present consecutive video frames and corresponding audio based saliency maps in the first and second rows respectively. NFoVs in multiple time steps that are used for 360\degree~content exploration are shown in the last row. Results show that audio based saliency maps can be effectively used for camera view panning in 360\degree~videos.}
\label{fig:360_NFoV_supp}
\end{figure*}

We use our per-frame saliency maps to compute the NFoV
tracks. 
Results in~\Fref{fig:360_NFoV_supp} show that our method capture salient viewpoints by using audio and vision for panning camera views in 360\degree~videos successfully. 

\end{document}